\documentclass[letterpaper]{article} 
\usepackage{aaai2026}  
\usepackage{times}  
\usepackage{helvet}  
\usepackage{courier}  
\usepackage[hyphens]{url}  
\usepackage{graphicx} 
\usepackage{natbib}  
\usepackage{caption} 
\usepackage{algorithm}
\usepackage{algorithmic}
\usepackage{amssymb}
\usepackage{amsmath}
\usepackage{xspace}
\usepackage{color}
\usepackage{multirow}
\usepackage{booktabs}
\usepackage{makecell}
\usepackage{multicol}
\usepackage{url}
\newcommand{\ourmethod}{{DMbaGCN}\xspace}

\usepackage{newfloat}
\usepackage{listings}
\DeclareCaptionStyle{ruled}{labelfont=normalfont,labelsep=colon,strut=off} 
\floatstyle{ruled}
\newfloat{listing}{tb}{lst}{}
\floatname{listing}{Listing}
\title{Dual Mamba for Node-Specific Representation Learning:\\  Tackling Over-Smoothing with Selective State Space Modeling}

\author{
    Xin He\textsuperscript{\rm 1},
    Yili Wang\textsuperscript{\rm 1},
    Yiwei Dai\textsuperscript{\rm 1},
    Xin Wang\textsuperscript{\rm 1}\thanks{Corresponding author}
}
\affiliations{
    \textsuperscript{\rm 1}School of Artificial Intelligence Jilin University\\

    hex23@mails.jlu.edu.cn, wangyili@jlu.edu.cn, daiyw25@mails.jlu.edu.cn,xinwang@jlu.edu.cn
}

\begin{document}

\maketitle

\begin{abstract}
Over-smoothing remains a fundamental challenge in deep Graph Neural Networks (GNNs), where repeated message passing causes node representations to become indistinguishable. While existing solutions, such as residual connections and skip layers, alleviate this issue to some extent, they fail to explicitly model how node representations evolve in a node-specific and progressive manner across layers.
Moreover, these methods do not take global information into account, which is also crucial for mitigating the over-smoothing problem. 
To address the aforementioned issues, in this work, we propose a Dual Mamba-enhanced Graph Convolutional Network (\ourmethod), which is a novel framework that integrates Mamba into GNNs to address over-smoothing from both local and global perspectives. \ourmethod consists of two modules: the Local State-Evolution Mamba (LSEMba) for local neighborhood aggregation and utilizing Mamba's selective state space modeling to capture node-specific representation dynamics across layers, and the Global Context-Aware Mamba (GCAMba) that leverages Mamba’s global attention capabilities to incorporate global context for each node. By combining these components, \ourmethod enhances node discriminability in deep GNNs, thereby mitigating over-smoothing. Extensive experiments on multiple benchmarks demonstrate the effectiveness and efficiency of our method.
\end{abstract}

\begin{links}
    \link{Code}{https://github.com/hexin5515/DMbaGCN}
    \link{Datasets}{https://github.com/hexin5515/DMbaGCN}
\end{links}

\section{Introduction}

Graph Neural Networks (GNNs) have become a dominant framework for learning on graph-structured data, achieving impressive performance across domains such as social networks~\cite{yang2024graph}, recommendation systems~\cite{khan2025heterogeneous}, molecular property prediction~\cite{sypetkowski2024scalability}, and knowledge graph reasoning~\cite{wang2023vqa}.
A key strength of most GNNs lies in the message-passing paradigm~\cite{miao2024rethinking,wang2025adagcl+}, where each node updates its representation by aggregating information from its neighbors. 
While message passing enables GNNs to aggregate local information effectively, it lacks an explicit mechanism for modeling how node representations evolve across layers in a progressive and node-specific manner. This is a key factor behind the \textbf{over-smoothing}~\cite{roth2024rank} phenomenon in GNNs.

\begin{figure}[t!]
  \includegraphics[width=1\linewidth]{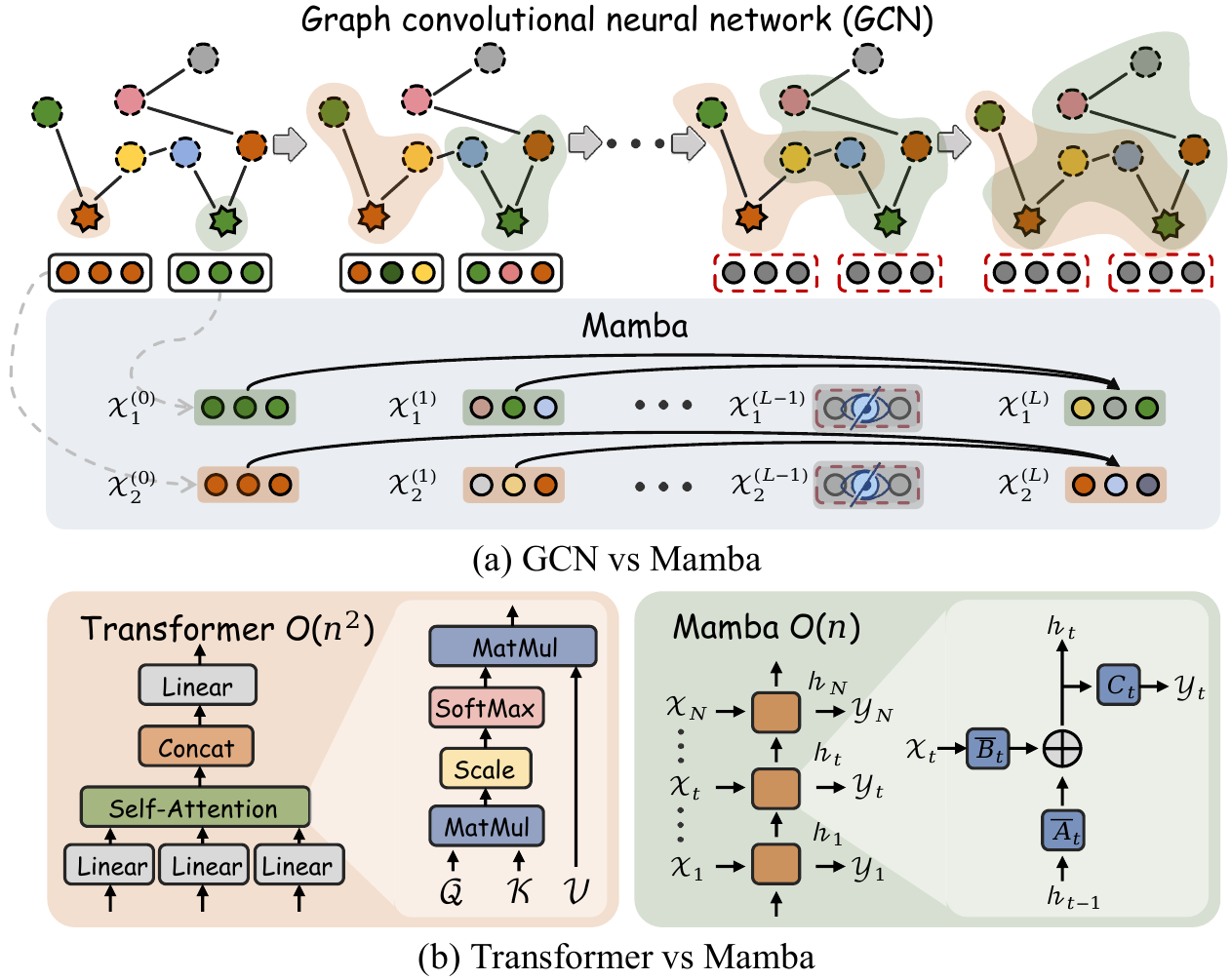}
  \caption{Comparison between GCN and Mamba (top), and comparison between Transformer and Mamba (bottom).}
  \label{fig:motivation}
\end{figure}
To tackle this problem, we seek a mechanism that can explicitly model how each node’s representation evolves with GNN depth, capturing both its layer-wise progression and node-specific dynamics.
Inspired by recent advances in state space modeling, we consider \textit{Mamba}, which captures long-range dependencies through input-dependent recurrence and has shown success in NLP~\cite{waleffe2024empirical} and CV~\cite{wang2025mamba} tasks. 
Unlike standard GNNs~\cite{velivckovic2023everything,GRASS}, Mamba inherently supports selective, position-aware modeling of sequential inputs, making it well-suited to guide how node features change layer by layer.
As illustrated in Figure~\ref{fig:motivation}(a), this property allows Mamba to adaptively shape each node’s representation trajectory, thereby mitigating over-smoothing.
This motivates an important question: \textbf{\textit{How can we apply Mamba to deep GNNs for effectively modeling the layer-wise evolution of node representations?}}

Even though Mamba effectively captures the layer-wise evolution of node representations, its modeling remains confined to local receptive fields and lacks access to global context~\cite{FuDHL0C25}.
Existing solutions often rely on Transformer-based models~\cite{hatamizadeh2025mambavision,wang2025vggt} to introduce global attention as they allow nodes to aggregate long-range information directly.
However, recent studies~\cite{ali2024hidden,gao2024matten} have demonstrated that Mamba can emulate global attention mechanisms similar to those used in Transformers.
More importantly, as shown in Figure~\ref{fig:motivation}(b), Mamba achieves this global modeling capability with significantly lower time complexity compared to Transformers.
This motivates another important question: \textbf{\textit{How can we leverage Mamba’s global modeling capability to efficiently inject global context into deep GNNs?}}

To address these issues, we propose a Dual Mamba-enhanced Graph Convolutional Network (\ourmethod), which introduces two key modules: The \textbf{Local State-Evolution Mamba} (LSEMba) and \textbf{Global Context-Aware Mamba} (GCAMba). LSEMba integrates multi-hop message passing with Mamba’s selective state space mechanism to track how node representations evolve across layers, while preserving local neighborhood information. This enables LSEMba to address the first challenge of capturing node-specific representation evolution across layers, effectively mitigating over-smoothing by preventing nodes from becoming indistinguishable as the depth increases.

On the other hand, GCAMba introduces a bidirectional Mamba model to capture global dependencies across the graph efficiently. Unlike standard GNNs that primarily focus on local neighborhood aggregation, GCAMba uses a bidirectional approach to ensure that each node can aggregate information from all other nodes, providing richer global context. This allows GCAMba to address the second challenge: efficiently capturing global dependencies without incurring the high computational cost of attention-based methods like Transformers. GCAMba’s bidirectional structure and linear time complexity $\mathcal{O}(N)$ make it an effective and efficient solution for global context aggregation in deep GNNs.

By combining LSEMba and GCAMba, \ourmethod offers a unified solution that effectively handles both node-specific evolution and global context aggregation. The two modules work together to provide both localized and global information to each node, ensuring that node representations evolve progressively and capturing long-range global dependencies efficiently for mitigating over-smoothing in GNNs.

The contributions of this work are summarized as follows:

\begin{itemize}
    \item We propose \ourmethod, a novel framework that integrates Mamba with GNNs to tackle the over-smoothing problem from both local and global perspectives.
    \item To address the challenge of modeling the layer-wise evolution of node representations, \ourmethod introduces a Local State-Evolution Mamba (LSEMba), which sequentially captures neighborhood information and models node representation evolution.
    \item For capturing global dependencies efficiently, \ourmethod incorporates the Global Context-Aware Mamba (GCAMba), which leverages global attention mechanism with linear computational complexity.
    \item Extensive experiments on multiple benchmark datasets demonstrate that \ourmethod achieves competitive performance and improves computational efficiency.
\end{itemize}
\section{Related Work}
\subsection{Over-smoothing in GNNs}
Over-smoothing occurs when repeated message passing makes node representations indistinguishable~\cite{li2018deeper}, leading to performance drops in tasks requiring fine-grained distinction~\cite{liu2024enhancing}. Prior works address this by decoupling propagation and transformation~\cite{chen2020simple}, limiting neighborhood ranges~\cite{wu2023demystifying,shen2024optimizing}, or randomly dropping edges~\cite{chen2025adedgedrop}. Attention-based methods also help by focusing on important neighbors~\cite{velivckovic2017graph}. 



\subsection{Graph Transformer}
Graph Transformers have gained increasing attention for their ability to capture long-range dependencies through global attention mechanisms~\cite{choi2024topology}. Unlike traditional GNNs~\cite{wang2024unifying,shen2025raising} that rely on local message passing~\cite{shen2025understanding}, Graph Transformers compute pairwise interactions between all nodes, making them well-suited for tasks requiring global context. This global view also helps alleviate over-smoothing, as each node can attend to distant yet relevant nodes without relying on deep propagation~\cite{lin2024gramformer}. Recent works further improve efficiency by introducing sparse attention patterns or structural encoding to adapt Transformers to graph data~\cite{chen2024sigformer,fu2024vcr}.



\subsection{Graph Mamba}
Mamba~\cite{han2024demystify,patro2024simba} was originally designed for sequence modeling, offering both selective information filtering and efficient global dependency modeling through a linear-time recurrence mechanism. Although its application to graph learning is still emerging, recent works have explored combining Mamba with GCNs to enhance long-range interaction modeling~\cite{behrouz2024graph,ding2024recurrent} and track over-smoothing~\cite{he2025mamba}. 
However, these works primarily leverage Mamba's ability to model dependencies in long sequences with selective state space.
In this work, we propose a unified framework that integrates Mamba into the GNNs, leveraging its selective filtering to preserve node-level distinction and its global modeling capacity to enhance representation expressiveness in deep networks.
\begin{figure*}[ht!]
  \includegraphics[width=\textwidth]{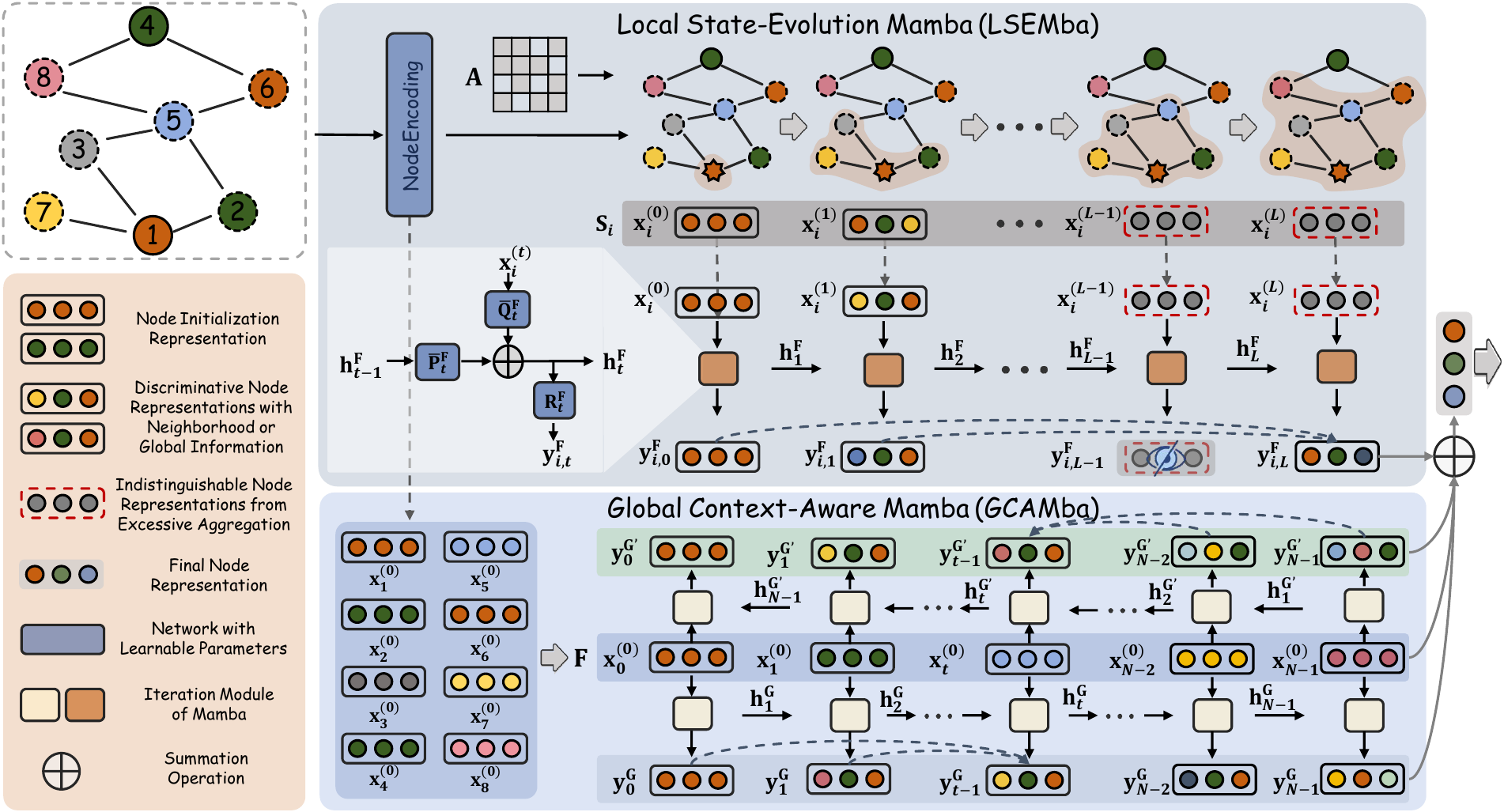}
  \caption{The Framework of \ourmethod. LSEMba models the evolution of node representations across GNN layers using Mamba's selective state space modeling. GCAMba aggregates global information for each node through bidirectional Mamba.}
  \label{fig:framework}
\end{figure*}

\section{Preliminary}
\subsection{Problem Statement}

Given an undirected graph \( \mathcal{G} = (\mathcal{V}, \mathcal{E}) \) with \( N \) nodes and \( M \) edges, the adjacency matrix \( \mathbf{A} \in \mathbb{R}^{N \times N} \) is defined such that \( \mathbf{A}_{ij} = 1 \) if there is an edge between node \( v_i \) and node \( v_j \), and \( \mathbf{A}_{ij} = 0 \) otherwise. Let \( \mathbf{X} \in \mathbb{R}^{N \times d} \) represent the node feature matrix, where each node \( v_i \) has a corresponding \( d \)-dimensional feature vector.

In node classification tasks, the goal is to predict node labels based on their features and the graph structure. Over-smoothing occurs as the depth of GNNs increases, causing node representations to converge and become indistinguishable. This phenomenon can be formalized as:
\begin{align}
\lim_{L \to \infty} \frac{1}{N} \sum_{i=1}^N \|\mathbf{h}_i^{(L)} - \mathbf{h}_j^{(L)}\|_2 \approx 0, \quad \forall i, j \in \mathcal{V}
\end{align}
where \( \mathbf{h}_i^{(L)} \) and \( \mathbf{h}_j^{(L)} \) are node representations at layer \( L \).

\subsection{General Graph Mamba}
Graph Mamba Networks (GMNs)~\cite{song2024breaking, wang2024graph} models the sequence of nodes in a graph iteratively, where each node's representation evolves not only based on its features but also the representations of all preceding nodes in the graph. 
This iterative process allows GMNs to capture long-range dependencies across the graph, while maintaining linear time complexity.
At each iteration, the representation of node is updated by considering all preceding nodes in the sequence,
which can be formulated as:
\begin{align}
    \mathbf{h}_{t}' &= \mathbf{P} \mathbf{h}_t + \mathbf{Q} \mathbf{x}_t, \\
    \mathbf{y}_t &= \mathbf{R} \mathbf{h}_t,
\end{align}
where $\mathbf{h}_t \in \mathbb{R}^N$ represents the hidden state of the node, $\mathbf{x}_t \in \mathbb{R}^N$ is the input node representations, and $\mathbf{P} \in \mathbb{R}^{N \times N}$, $\mathbf{Q} \in \mathbb{R}^N$, $\mathbf{R} \in \mathbb{R}^N$ are parameters generated by small neural networks that depend on the node-specific initial representations. 
This means that each position generates its transition parameters, endowing Mamba with an input-dependent attention mechanism while still keeping linear computational complexity.
To enable efficient training, the continuous-time model is discretized to improve computational efficiency:
\begin{align}
\bar{\mathbf{P}} &= \exp(\Delta \cdot \mathbf{P}),  \\
\bar{\mathbf{Q}} &= (\Delta \cdot \mathbf{P})^{-1} \left( \exp(\Delta \cdot \mathbf{P}) - \mathbf{I} \right) \cdot \Delta \mathbf{Q},
\end{align}
where $\Delta$ is a learned discretization step size. This results in a discrete-time update rule:
\begin{align}
\mathbf{h}_t &= \bar{\mathbf{P}} \mathbf{h}_{t-1} + \bar{\mathbf{Q}} \mathbf{x}_t, \\
\mathbf{y}_t &= \mathbf{R} \mathbf{h}_t.
\end{align}

The above selective state space model is equivalent to a convolution operation~\cite{dao2024transformers}, where the kernel $\mathbf{K} = (\mathbf{R}\bar{\mathbf{Q}}, \mathbf{R}\bar{\mathbf{P}}\bar{\mathbf{Q}}, \dots, \mathbf{R}\bar{\mathbf{P}}^{L-1}\bar{\mathbf{Q}})$ is computed efficiently in $\mathcal{O}(N)$. 
This contrasts with the global attention mechanism in Graph Transformers, which attends to all nodes at once, causing quadratic complexity $\mathcal{O}(N^2)$.
In this work, we leverage Mamba's selective state space modeling and global attention mechanism to enhance node discriminability, allowing the model to learn from both local and global context, thereby significantly reducing over-smoothing.

\section{DMbaGCN}

The framework of \ourmethod, shown in Figure~\ref{fig:framework}, integrates two key modules: LSEMba (Local State-Evolution Mamba) and GCAMba (Global Context-Aware Mamba). These Mamba-based modules are specifically designed to handle both the progressive evolution of node representations and efficient aggregation of global dependencies.

Detailed descriptions of each module and its role in the framework are provided in the following sections.

\subsection{Local State-Evolution Mamba (LSEMba)}
LSEMba integrates two key components: \textbf{local neighborhood aggregation} and \textbf{adaptive representation evolution}. First, it aggregates information from a node’s local neighborhood through stacked graph convolution layers, capturing high-order neighborhood features. Then, it models the progressive evolution of node representations across layers using Mamba’s selective state space modeling, which enables each node's representation to adapt layer by layer. This combined approach allows LSEMba to maintain node-specific dynamics while preserving local features, effectively addressing the over-smoothing problem in GNNs.

The above process begins with neighborhood aggregation~\cite{BQN}, where each node gathers multi-hop information from its local neighborhood. This multi-scale local context then serves as the input for the representation evolution process, allowing LSEMba to track and adapt node representations across layers, ensuring node discriminability as depth increases. The node representation at depth \(l\) is computed as:
\begin{align}
 \label{eq:mal}
    \textbf{X}^{(l)} = \Tilde {\mathbf{D}}^{-\frac{1}{2} } \mathbf{A} \Tilde{\mathbf{D}}^{-\frac{1}{2}} \mathbf{X}^{(l-1)},
\end{align}
where $\Tilde{\mathbf{D}}$ is the diagonal degree matrix, and $\mathbf{A}$ is the adjacency matrix. $\mathbf{X}^{(l-1)}$ and $\mathbf{X}^{(l)}$ refer to the node representations after aggregating information from the $(l-1)$-hop and $l$-hop neighborhoods, respectively.

These node representations across layers are then ordered by ascending depth to form a sequence. For node \(v_i\), the sequence of its representations is:
\begin{align}
 \label{eq:seq}
    \textbf{S}_{i} = [\mathbf{x}_i^{(0)},\mathbf{x}_i^{(1)},...,\mathbf{x}_i^{(L)}],
\end{align}
where $\mathbf{S}_{i}$ is the sequence of node representations for node $v_i$, $\mathbf{x}_i^{(l)}$ is the representation of node $v_i$ at layer $l$, with $l$ ranging from 0 to $L$. $L$ is the total number of layers in the GNNs. 
These sequences are then modeled by Mamba using its selective state space mechanism, which recursively models how each node's representation adaptive evolves across layers while preserving node-specific features.

To model the adaptive evolution of node representations across layers, LSEMba utilizes Mamba’s state space mechanism. The first step involves initializing the Mamba parameters $\mathbf{Q}^{\mathbf{F}}$, $\mathbf{R}^{\mathbf{F}}$, and learnable step size $\mathbf{\Delta}^{\mathbf{F}}$ based on the sequence of node representations $\mathbf{S}$, which is formulated as:
\begin{equation} \label{eq:initial}
\mathbf{Q}^{\mathbf{F}}=f_{\theta_1}(\mathbf{S}), \mathbf{R}^{\mathbf{F}}=f_{\theta_2}(\mathbf{S}),
\mathbf{\Delta}^{\mathbf{F}}=f_{\theta_3}(\mathbf{S}),
\end{equation}
where $\theta_1$, $\theta_2$ and $\theta_3$ are learnable parameters. These parameters govern the selective state evolution and control how node features change layer by layer.

Next, the initialized parameters $\mathbf{P}^{\mathbf{F}}$ and $\mathbf{Q}^{\mathbf{F}}$ undergo discretization to enable efficient training and optimization\footnote{We apply the HiPPO-Legs initialization~\cite{gu2020hippo} to the parameter $\mathbf{P}^{\mathbf{F}}$ to enable LSEMba handle longer contexts.}:
\begin{align}
    \bar{\mathbf{P}}^{\mathbf{F}}&={\rm exp}(\mathbf{\Delta}^{\mathbf{F}} \mathbf{P}^{\mathbf{F}}),\\
    \bar{\mathbf{Q}}^{\mathbf{F}}&=(\mathbf{\Delta}^{\mathbf{F}} \mathbf{P}^{\mathbf{F}})^{-1}({\rm exp}(\mathbf{\Delta}^{\mathbf{F}} \mathbf{P}^{\mathbf{F}})-\mathbf{I})\cdot \mathbf{\Delta} ^{\mathbf{F}}\mathbf{Q}^{\mathbf{F}},
\end{align}
where $\bar{\mathbf{P}}^{\mathbf{F}}$ and $\bar{\mathbf{Q}}^{\mathbf{F}}$ represent the discretized parameters. This discretization enables efficient computation during training, ensuring that LSEMba can model long-range dependencies and adaptively track node evolution across layers without sacrificing computational efficiency.

To integrate multi-hop neighborhood information and generate expressive node representations for downstream tasks, we iterate through the sequence of neighborhood information for each node. This process starts from shallow layers and progresses to deeper layers, using the initialized model parameters.
The parameter $\bar{\mathbf{P}}^{\mathbf{F}}_t$ controls the retention of low-order neighborhood information at each step, $\bar{\mathbf{Q}}^{\mathbf{F}}_t$  determines the amount of corresponding-order neighborhood information injected into the node features, and  $\mathbf{R}^{\mathbf{F}}_t$ generates the output representation for downstream tasks. The iterative process is formulated as:
\begin{align} \label{eq:sql_1}
\mathbf{h}_{t}^{\mathbf{F}}=\bar{\mathbf{P}}_t^{\mathbf{F}}\cdot&\mathbf{h}_{t-1}^{\mathbf{F}}+\bar{\mathbf{Q}}_t^{\mathbf{F}}\cdot\mathbf{x}^{(t)}, \\
\mathbf{y}_{i,t}^{\mathbf{F}}&=\mathbf{R}_t^{\mathbf{F}}\cdot\mathbf{h}_{t}^{\mathbf{F}},
\end{align}
where $\mathbf{x}_t$ is the node representation that includes the $t$-hop neighborhood information, $\mathbf{h}_t^{\mathbf{F}}$ is the hidden state after integrating the $t$-hop neighborhood information, and $\mathbf{y}_{i,t}^{\mathbf{F}}$ is the output representation for node $v_i$.


LSEMba effectively integrates multi-hop neighborhood information and adapts node representations through local aggregation. While it successfully mitigates over-smoothing by preserving node-specific feature evolution, it is limited in its ability to capture global dependencies across the graph. This focus on local context restricts its capacity to maintain distinct node features, which are essential for tackling over-smoothing in deeper GNNs. To address this gap, we introduce the Global Context-Aware Mamba (GCAMba).

\subsection{Global Context-Aware Mamba (GCAMba)}
GCAMba addresses the limitations of local aggregation by enabling the model to capture \textbf{global dependencies} across the entire graph. By adopting a bidirectional Mamba model, GCAMba allows information to flow in both directions through the node sequence. This bidirectional approach ensures that each node aggregates global context, enriching its representation with long-range dependencies. As a result, GCAMba provides a more complete and expressive node representation, complementing LSEMba by overcoming the constraints of local neighborhood aggregation.

To capture global dependencies across the entire graph, GCAMba processes the node sequence, which is constructed by concatenating the initial representations of all nodes. Specifically, the sequence is defined as:
\begin{align}
    \mathbf{F}=[\mathbf{x}_1^{(0)},\mathbf{x}_2^{(0)},\cdots,\mathbf{x}_N^{(0)}],
\end{align}
where $\mathbf{F}$ is the input sequence, constructed by concatenating the initial representations of all nodes. $\mathbf{x}_i^{(0)}$ denotes the original representation of node $v_i$, and $N$ is the number of nodes in the graph. We initialize the Mamba parameters $\bar{\mathbf{P}}^{\mathbf{G}}$, $\bar{\mathbf{Q}}^{\mathbf{G}}$, $\mathbf{R}^{\mathbf{G}}$, and $\mathbf{\Delta}^{\mathbf{G}}$ in GCAMba using the same strategy as in LSEMba, and omit the detailed formulation here for brevity.

To enhance node discriminability, GCAMba iteratively processes the node representation sequence. This iterative process is formulated as follows: 
\begin{align}
\mathbf{h}_1^{\mathbf{G}}=\bar{\mathbf{Q}}^{\mathbf{G}}_1\mathbf{x}_1^{(0)},
&\mathbf{h}_2^{\mathbf{G}}=\bar{\mathbf{P}}^{\mathbf{G}}_2\bar{\mathbf{Q}}^{\mathbf{G}}_1\mathbf{x}_1^{(0)}+\bar{\mathbf{Q}}^{\mathbf{G}}_2\mathbf{x}_2^{(0)},\\
\mathbf{y}_1^{\mathbf{G}}=\mathbf{R}^{\mathbf{G}}_1\bar{\mathbf{Q}}^{\mathbf{G}}_1\mathbf{x}_1^{(0)},
&\mathbf{y}_2^{\mathbf{G}}=\mathbf{R}^{\mathbf{G}}_2\bar{\mathbf{P}}^{\mathbf{G}}_2\bar{\mathbf{Q}}^{\mathbf{G}}_1\mathbf{x}_1^{(0)}+\mathbf{R}^{\mathbf{G}}_2\bar{\mathbf{Q}}^{\mathbf{G}}_2\mathbf{x}_2^{(0)},
\end{align}
continuing this process, the general form of the iterative process can be expressed as follows:
\begin{align}
\mathbf{h}_t^{\mathbf{G}}=\sum_{j=1}^t(\Pi_{k=j+1}^t \bar{\mathbf{P}}^{\mathbf{G}}_k)\bar{\mathbf{Q}}^{\mathbf{G}}_j\mathbf{x}_j^{(0)}, \\
    \mathbf{y}_t^{\mathbf{G}} = \mathbf{R}^{\mathbf{G}}_t\sum_{j=1}^t(\Pi_{k=j+1}^t\bar{\mathbf{P}}_k^{\mathbf{G}})\bar{\mathbf{Q}}^{\mathbf{G}}_j\mathbf{x}^{(0)}_j,
\end{align}
where $\mathbf{h}_t^{\mathbf{G}}$ represents the hidden state at step $t$, aggregating information from all previous nodes in the sequence. $\mathbf{x}_j^{(0)}$ denotes the initial representation of node $v_j$. $\mathbf{y}_t^{\mathbf{G}}$ is the representation of node $v_t$ after aggregating global context information across all previous nodes in the sequence. 

Unidirectional Mamba only captures dependencies between the current node and its preceding nodes, which limits its ability to model global relationships that involve future nodes in the sequence. To address this limitation, GCAMba uses a bidirectional Mamba model, allowing information to flow in both directions through the node sequence. This enables each node to aggregate global context from all other nodes in the graph, resulting in more comprehensive node representations that enhance node discriminability.
The modeling process is defined as follows:
\begin{align}\label{eq:global_rep}
    \hat{\mathbf{Y}}^{\mathbf{G}}=(1-\beta)(f_{\varphi}(\mathbf{F}) + {\rm Re}(f_{\varphi}({\rm Re}(\mathbf{F})))) + \beta \mathbf{X}^{(0)},
\end{align}
where $\varphi$ is learnable parameters of the Mamba model $f_{\varphi}(\cdot)$ in GCAMba, constructed similarly to those in the LSEMba. ${\rm Re}(\cdot)$ denotes the operation that reverses the input sequence. $\mathbf{F}$  is the initial representation matrix for all nodes, and $\hat{\mathbf{Y}}^{\mathbf{G}}$ represents the final output after aggregating complete global context from both directions.
The hyperparameter $\beta$ denotes the hyperparameter for the residual connection.

\begin{algorithm}[!b]
    \caption{DMbaGCN}\label{algor:dmabgcn}
    \raggedright
    \textbf{Input}: Adjacency matrix $\mathbf{A} \in \mathbb{R}^{N \times N}$, feature matrix $\mathbf{X}\in \mathbb{R}^{N\times d}$, state matrixs $\mathbf{P}^{\mathbf{F}}$, $\mathbf{P}^{\mathbf{G}}$ , learnable parameters $\theta=\{\theta_1, \theta_2, \theta_3\}, \varphi=\{\varphi_1, \varphi_2, \varphi_3\}$. \\
    \textbf{Output}: The updated node representations $\mathbf{Z}$.
    
    \begin{algorithmic}[1] 
    \WHILE{not convergent}
    \STATE Compute $\mathbf{X}^{(1)}, \mathbf{X}^{(2)}, \cdots, \mathbf{X}^{(L)}$, via Eq.\ref{eq:mal};
    \STATE Compute $\mathbf{Q}^{\mathbf{F}(\mathbf{G})}$, $\mathbf{R}^{\mathbf{F}(\mathbf{G})}$ and $\mathbf{\Delta}^{\mathbf{F}(\mathbf{G})}$ via Eq.\ref{eq:initial};
    \STATE Compute $\mathbf{Y}^{\mathbf{F}}_L$ via Eq.13 and Eq.14;
    \STATE Compute $\hat{\mathbf{Y}}^{\mathbf{G}}$ via Eq.\ref{eq:global_rep};
    \STATE Compute $\mathbf{Z}$ via Eq.\ref{eq:final_rep};
    \STATE Update learnable parameters via back propagation;
    \ENDWHILE
    \RETURN Final node representations $\mathbf{Z}$.
    \end{algorithmic}
\end{algorithm}

Finally, \ourmethod adopts a simple yet effective strategy to integrate the node representations from LSEMba and GCAMba, producing the final node representations $\mathbf{Z}$ used for downstream tasks, which is formulated as follows:
\begin{align}\label{eq:final_rep}
    \mathbf{Z}=\alpha\mathbf{Y}^{\mathbf{F}}_{L} + (1-\alpha)\hat{\mathbf{Y}}^{\mathbf{G}},
\end{align}
where $\mathbf{Y}^{\mathbf{F}}_L$ denotes the output from LSEMba, which captures the layer-wise evolution of node representations based on local neighborhoods. $\hat{\mathbf{Y}}^{\mathbf{G}}$ denotes the output from GCAMba, encoding comprehensive global contextual information for each node. 
$\alpha$ is a hyperparameter that controls the contribution of each component.
These two representations enhance the discriminability of node representations from both local and global perspectives, jointly alleviating the over-smoothing problem in GNNs. The detailed implementation of \ourmethod is shown in Algorithm~\ref{algor:dmabgcn}.
\section{Experiment}

\begin{table*}[t]
\centering
\scalebox{0.90}{
\begin{tabular}{c|c|cc|cc|cc}
\toprule
\multicolumn{1}{c}{\textbf{}}&\multicolumn{1}{c|}{\textbf{}}  & \multicolumn{1}{c}{\textbf{CoraFull}}& \multicolumn{1}{c}{\textbf{Pubmed}}& \multicolumn{1}{|c}{\textbf{Computers}}& \multicolumn{1}{c|}{\textbf{Photo}}& \multicolumn{1}{c}{\textbf{CS}}& \multicolumn{1}{c}{\textbf{Physics}}\\
\midrule
\multirow{3}{*}{\makecell[c]{\textbf{GNN}}}
&\multirow{1}{*}{\textbf{GCN}}  &   70.69 ± 0.37 &   87.82 ± 0.30&  91.07 ± 0.39 &   93.77 ± 0.31   &    93.75 ± 0.26 & 96.34 ± 0.21 \\
&\multirow{1}{*}{\textbf{GAT}}
&  71.42 ± 0.15 &   88.72 ± 0.17
 &   90.94 ± 0.18 &  93.81 ± 0.19 &   93.84 ± 0.11 & 96.43 ± 0.12 \\
&\multirow{1}{*}{\textbf{SGC}}
&   70.04 ± 0.25 &   87.91 ± 0.36 &    91.62 ± 0.35 &    93.61 ± 0.43 &  93.44 ± 0.18 &  96.43 ± 0.20 \\ \midrule
\multirow{4}{*}{\makecell[c]{\textbf{Deep GNN}}}&
                \multirow{1}{*}{\textbf{APPNP}}
&  69.37 ± 0.35 &   88.94 ± 0.31 &  89.51 ± 0.31   &   94.10 ± 0.35 &    \underline{95.65 ± 0.17}   & 97.01 ± 0.17   \\
               & \multirow{1}{*}{\textbf{GCNII}}
&  \underline{72.23 ± 0.50}&  \underline{90.15 ± 0.31} &  84.71 ± 0.40  &    92.46 ± 0.70 &  95.46 ± 0.14  &  \underline{97.09 ± 0.13} \\
&\multirow{1}{*}{\textbf{GPRGNN}}
&   71.16 ± 0.50  &   89.64 ± 0.40&  91.80 ± 0.35  &  94.97 ± 0.12 &  95.49 ± 0.19& 97.05 ± 0.13\\ 
&\multirow{1}{*}{\textbf{SSGC}}&   70.51 ± 0.25&   88.39 ± 0.41 &  \underline{91.98 ± 0.36} &  94.28 ± 0.33 & 93.99 ± 0.23 &  96.60 ± 0.15 \\ 
\midrule
\multirow{5}{*}{\textbf{\makecell[c]{Graph \\ Transformer}}}&
\multirow{1}{*}{\textbf{GT}}&   61.05 ± 0.38 &   88.79 ± 0.12 & 91.18 ± 0.17 &  94.74 ± 0.13  & 94.64 ± 0.13&  97.05 ± 0.05 \\ 
&\multirow{1}{*}{\textbf{Graphormer}}&   OOM &   OOM
 &  OOM &  92.74 ± 0.14  & 94.64 ± 0.13 &  OOM\\ 
&\multirow{1}{*}{\textbf{SAN}}&   59.01 ± 0.34 &   88.22 ± 0.15 &  89.93 ± 0.16 &  94.86 ± 0.10  & 94.51 ± 0.15 &  OOM\\ 
&\multirow{1}{*}{\textbf{GraphGPS}}&   55.76 ± 0.23 &  88.94 ± 0.16 &  OOM &  95.06 ± 0.13  & 93.93 ± 0.15 &  OOM \\ 
&\multirow{1}{*}{\textbf{Spexphormer}}& 71.84 ± 0.25  &  89.87 ± 0.19  &  91.09 ± 0.08 &  \underline{95.33 ± 0.49} & 95.00 ± 0.15 &  96.70 ± 0.05 \\ 
\midrule
\multirow{6}{*}{\textbf{\makecell[c]{Graph \\Mamba}}}&\multirow{1}{*}{\textbf{MbaGCN}}&  71.76 ± 0.31  & 89.32 ± 0.24  & 90.39 ± 0.21  &  94.41 ± 0.75  & 95.33 ± 0.12 &  96.64 ± 0.08  \\ 
\cmidrule{2-8}
&\multirow{2}{*}{\textbf{\makecell[c]{\ourmethod \\ 
w/o GCAMba}}}
&   \multirow{2}{*}{71.51 ± 0.37} &   \multirow{2}{*}{89.67 ± 0.47}&  \multirow{2}{*}{91.74 ± 0.34}    &  \multirow{2}{*}{94.44 ± 0.30} &   \multirow{2}{*}{95.32 ± 0.15}& \multirow{2}{*}{96.37 ± 0.16} \\ 
&\multicolumn{1}{c|}{}&&\multicolumn{1}{c|}{}&&\multicolumn{1}{c|}{}\\\cmidrule{2-8}
&\multirow{2}{*}{\textbf{\makecell[c]{\ourmethod \\ 
w/o LSEMba}}}
&   \multirow{2}{*}{62.15 ± 0.52} &   \multirow{2}{*}{88.14 ± 0.39}&  \multirow{2}{*}{84.18 ± 0.47}    &  \multirow{2}{*}{89.99 ± 0.50} &   \multirow{2}{*}{93.71 ± 0.23}& \multirow{2}{*}{95.71 ± 0.14} \\ 
&\multicolumn{1}{c|}{}&&\multicolumn{1}{c|}{}&&\multicolumn{1}{c|}{}\\\cmidrule{2-8}
&\multirow{1}{*}{\textbf{\ourmethod (ours)}}
&   \textbf{72.26 ± 0.20} &   \textbf{90.49 ± 0.33}&  \textbf{92.49 ± 0.37}    &  \textbf{95.61 ± 0.18} &   \textbf{96.00 ± 0.21} & \textbf{97.13 ± 0.14}\\
\bottomrule
\end{tabular}}
\caption{Summary of classification accuracy (\%). The best result for each benchmark is highlighted in \textbf{bold}, and the second-best result is emphasized with an \underline{underline}.} \label{tab:performance_com}
\end{table*}
To validate the effectiveness and efficiency of \ourmethod, we conduct comprehensive experiments on several benchmark datasets. aiming to answer three key questions: 
\begin{itemize}
    \item \textbf{Q1}: Does \ourmethod outperform baseline models across various datasets?
    \item \textbf{Q2}: Can \ourmethod effectively mitigate the over-smoothing problem in deep GNNs? 
    \item \textbf{Q3}: Is \ourmethod more efficient than previous global attention-based models for capturing global information? 
\end{itemize}
\subsection{Experimental Setting}
\noindent$\rhd$ \textbf{Dataset}: 
We evaluate our proposed method on the widely used benchmark datasets from different domains, including two citation graphs (\textbf{CoraFull}, \textbf{Pubmed})~\cite{shen2024graph}, two web graphs (\textbf{Computers}, \textbf{Photo})~\cite{chen2024leveraging}, and two co-authorship graphs (\textbf{CS}, \textbf{Physics})~\cite{fu2024vcr}. These datasets provide a diverse yet consistent evaluation setting for node classification under fully supervised scenarios. 
For each dataset, we use the publicly available node features and labels. Following prior work~\cite{fu2024vcr}, we randomly split the nodes into training, validation, and test sets with a fixed ratio of 60\%/20\%/20\%. All results are averaged over 10 random splits to ensure robustness. The detailed of each dataset can be found in the Appendix A.1.

\noindent$\rhd$ \textbf{Baselines}: 
To validate the effectiveness and efficiency of \ourmethod, we compare \ourmethod with several kinds of GNN models, including classical models \textbf{GCN}~\cite{kipf2016semi}, \textbf{GAT}~\cite{velivckovic2017graph}, and \textbf{SGC}~\cite{wu2019simplifying}, deep GNN models \textbf{APPNP}~\cite{gasteiger2018predict}, \textbf{GCNII}~\cite{chen2020simple}, \textbf{GPRGNN}~\cite{chien2020adaptive}, \textbf{SSGC}~\cite{zhu2021simple}, graph transformer models \textbf{GT}~\cite{dwivedi2020generalization}, \textbf{Graphormer}~\cite{ying2021transformers}, \textbf{SAN}~\cite{kreuzer2021rethinking}, \textbf{GraphGPS}~\cite{rampavsek2022recipe}, \textbf{Spexphormer}~\cite{shirzad2024even} and graph mamba model \textbf{MbaGCN}~\cite{he2025mamba}. Additional information of the baseline models is available in Appendix A.2.

\subsection{Performance Evaluation (Q1)}
We evaluate \ourmethod on six benchmark datasets against a broad range of baseline models. As shown in Table~\ref{tab:performance_com}, it consistently achieves competitive or superior performance. We analyze the results from three perspectives:

\noindent$\rhd$ \textbf{Performance against GNNs and Deep GNNs:}  
Compared to shallow GNNs (e.g., GCN, GAT, SGC), which suffer from limited receptive fields, and deep GNNs (e.g., GCNII, GPRGNN, SSGC), which rely on static propagation schemes, \ourmethod delivers better performance across all datasets—achieving 97.13\% on Physics and 96.00\% on CS. These gains come from the adaptive depth-wise modeling of LSEMba and the global integration of GCAMba, which together enhance representational power.

\noindent$\rhd$ \textbf{Performance against Graph Transformer:}
While Transformer-based models (e.g., Graphormer, GraphGPS) capture global dependencies, they incur high time and memory costs, and often fail on large graphs due to scalability issues. In contrast, \ourmethod maintains strong performance without such limitations, owing to GCAMba’s use of Mamba for efficient global modeling with linear complexity.

\noindent$\rhd$ \textbf{Performance against Graph Mamba:}
Compared to MbaGCN  which applies Mamba only locally, the dual Mamba integration strategy of \ourmethod leads to consistent improvements such as +1.17\% on Pubmed and +1.20\% on Photo. This confirms the benefit of jointly modeling node-specific evolution and global context.

\begin{table*}[t]
\renewcommand{\arraystretch}{1}
\centering
\scalebox{0.77}{
\begin{tabular}{c|ccccc|ccccc}
\toprule
\multicolumn{1}{c|}{\textbf{Layers}}  & \multicolumn{1}{c}{\textbf{2}}& \multicolumn{1}{c}{\textbf{4}}& \multicolumn{1}{c}{\textbf{8}}& \multicolumn{1}{c}{\textbf{16}}& \multicolumn{1}{c}{\textbf{32}}& \multicolumn{1}{|c}{\textbf{2}}& \multicolumn{1}{c}{\textbf{4}}& \multicolumn{1}{c}{\textbf{8}}& \multicolumn{1}{c}{\textbf{16}}& \multicolumn{1}{c}{\textbf{32}}\\
\midrule
\textbf{Dataset}&\multicolumn{5}{c|}{\textbf{Photo}}&\multicolumn{5}{c}{\textbf{Pubmed}} \\
\cmidrule{1-11}
\multirow{1}{*}{\textbf{GCN}}  &  \textbf{93.77±0.31}
 & \underline{91.39±0.26} & 85.97±0.71 & 43.63±0.48 & 24.03±0.56&  \textbf{87.82±0.30}
 & \underline{85.76±0.31} & 84.29±0.29 & 77.94±3.98 & 45.69±3.09\\
\multirow{1}{*}{\textbf{SGC}}& \textbf{93.61±0.43}  & \underline{92.07±0.56} & 89.76±0.27 & 86.33±0.54 & 79.19±0.90 & \textbf{87.91±0.36}  & \underline{85.91±0.30} & 84.47±0.38 & 83.46±0.23 & 82.34±0.33\\ \midrule
                \multirow{1}{*}{\textbf{APPNP}}&
94.05±0.39  & 93.95±0.41 & \textbf{94.10±0.35} & \underline{94.08±0.38} & 94.07±0.41 &
88.86±0.39  & \textbf{88.94±0.31} & 88.92±0.43 & 88.93±0.41 & \underline{88.94±0.36}\\
                \multirow{1}{*}{\textbf{GCNII}} & 91.96±0.49 & 91.46±0.65 & 91.93±0.58 & \underline{92.07±0.92}  & \textbf{92.46±0.70} & \underline{89.66±0.34}  & 89.56±0.32& 89.19±0.27 & 88.64±0.27 & \textbf{90.15±0.31}\\
\multirow{1}{*}{\textbf{GPRGNN}}
& 94.77±0.31  & 94.90±0.22 & \textbf{94.97±0.12} & \underline{94.90±0.11} &  94.77±0.28& 89.54±0.30  & 89.58±0.39 & \underline{89.61±0.27} & \textbf{89.64±0.40} & 89.45±0.32\\ 
\multirow{1}{*}{\textbf{SSGC}}& \underline{94.13±0.35}  & \textbf{94.28±0.33} & 94.03±0.39 & 93.60±0.37 & 93.16±0.29  & \underline{88.24±0.29} &\textbf{88.39±0.41} & 87.22±0.23  & 87.61±0.25 & 87.17±0.30\\  \cmidrule{1-11} 
\multirow{1}{*}{\textbf{MbaGCN}}& \textbf{94.77±0.52}& \underline{93.87±0.31} & 92.47±0.17 & 92.23±0.34 & 91.99±0.34  & \textbf{89.23±0.17} & \underline{89.18±0.42} & 89.13±0.39  & 89.16±0.33 & 89.14±0.45 \\ \cmidrule{1-11}
\multirow{2}{*}{\textbf{\makecell[c]{\ourmethod \\w/o GCAMba}}}& \multirow{2}{*}{94.42±0.41}  & \multirow{2}{*}{\textbf{94.44±0.30}}
& \multirow{2}{*}{94.24±0.20} & \multirow{2}{*}{94.28±0.27} & \multirow{2}{*}{\underline{94.43±0.29}}& \multirow{2}{*}{89.65±0.43}  & \multirow{2}{*}{89.61±0.37}
& \multirow{2}{*}{\underline{89.65±0.41}} & \multirow{2}{*}{89.43±0.37} & \multirow{2}{*}{\textbf{89.67±0.47}}\\ &&&&&\multicolumn{1}{c|}{} \\\cmidrule{2-11}
\multirow{2}{*}{\textbf{\makecell[c]{\ourmethod \\(ours)}}}& \multirow{2}{*}{94.59±0.33}  & \multirow{2}{*}{94.42±0.32}
& \multirow{2}{*}{95.35±0.24} & \multirow{2}{*}{\textbf{95.61±0.18}} & \multirow{2}{*}{\underline{95.46±0.37}}& \multirow{2}{*}{90.21±0.34}  & \multirow{2}{*}{90.29±0.36}
& \multirow{2}{*}{\textbf{90.49±0.33}} & \multirow{2}{*}{90.41±0.33} & \multirow{2}{*}{\underline{90.44±0.37}}\\ &&&&&\multicolumn{1}{c|}{} \\ 
\bottomrule
\end{tabular}}
\caption{Classification accuracy (\%) comparison under different layer configurations. The best result across different layer configurations is highlighted in \textbf{bold}, and the second-best result is emphasized with an \underline{underline}.} \label{tab:performance_com_under_dif_lay}
\end{table*}

\subsection{Layer-Wise Performance Trends (Q2)}
To assess whether \ourmethod effectively alleviates over-smoothing in deep GNNs, we conduct experiments with varying layer depths on the Photo and Pubmed datasets (Table~\ref{tab:performance_com_under_dif_lay}). 
The experimental results for the other four datasets
(CoraFull, Computer, CS and Physics) can be found in Appendix B.1.
We analyzes the results from two perspectives:

\noindent$\rhd$ \textbf{Performance against GNNs and Deep GNNs:}
Traditional GNNs like GCN and SGC suffer severe performance degradation as depth increases, due to indiscriminate neighbor aggregation that leads to over-smoothing. For instance, GCN's accuracy on Photo drops from 93.77\% (2 layers) to 24.03\% (32 layers). Deep GNNs such as GCNII and SSGC improve stability, but still rely on static propagation schemes that limit adaptability. For example, SSGC drops from 94.13\% to 93.16\%.
In contrast, \ourmethod remains consistently robust across depths, achieving 95.46\% on Photo and 90.44\% on Pubmed at 32 layers. This strong depth resilience stems from LSEMba’s ability to model node-specific representation evolution across layers, and GCAMba’s role in incorporating global context. 

\noindent$\rhd$ \textbf{Performance against Graph Mamba:}
The Graph Mamba model MbaGCN shows a slight performance drop with increasing depth, likely due to optimization challenges from its complex architecture (e.g., 94.77\% to 91.99\% on Photo and 89.23\% to 89.14\% on Pubmed).
In contrast, \ourmethod not only adopts a simpler structure but also incorporates global context through GCAMba, resulting in more stable performance (e.g., 94.59\% to 95.61\% on Photo and 90.12\% to 90.49\% on Pubmed) across depths and consistently competitive results under various layer configurations.

\subsection{Efficiency Analysis (Q3)}
We compare the efficiency of GCN, GT, Spexphormer, our method (\ourmethod) and its ablation without GCAMba in four datasets, evaluating memory and time costs. Memory and time performance are discussed separately.

\begin{figure}[t!]
\includegraphics[width=1\linewidth]{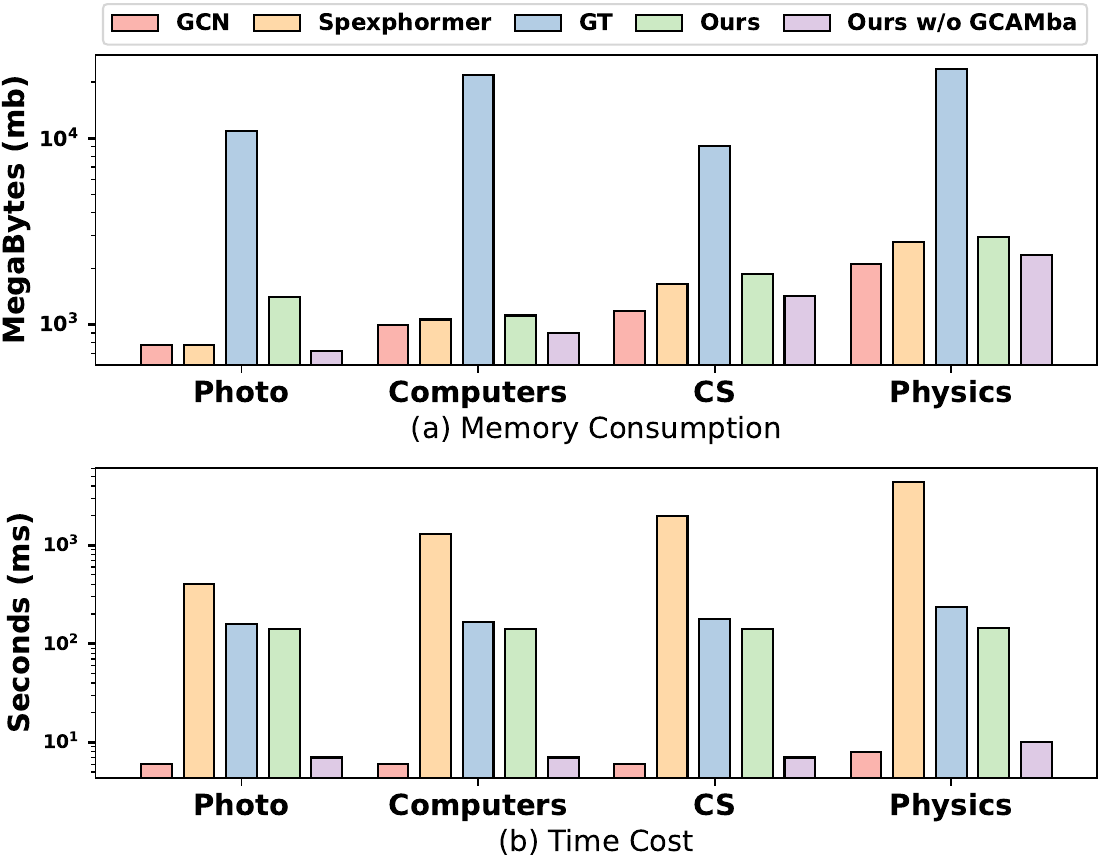}
  \caption{Comparison of Time and Memory Consumption.}
  \label{fig:time_memory_cost}
\end{figure}

\noindent$\rhd$ \textbf{Memory Consumption Analysis}: 
As shown in Figure~\ref{fig:time_memory_cost} (a), GCN consumes the least memory ($10^3$ MB) due to its simple structure. GT incurs the highest memory cost ($10^4$ MB) as it replaces GAT’s attention with a Transformer. Spexpormer reduces memory usage to $10^3$ MB through structural optimization. Our DMbaGCN achieves similarly low memory consumption, benefiting from Mamba’s efficient design. Its ablation variant matches GCN's memory cost, indicating that GCAMba accounts for most of the overhead.

\noindent$\rhd$ \textbf{Time Cost Analysis}:
As shown in Figure~\ref{fig:time_memory_cost}(b), although Spexpormer significantly reduces memory usage compared to GT, its additional sampling strategy leads to a higher time cost ($10^3$ ms). In contrast, GT avoids sampling and maintains a simpler design, keeping time consumption at the $10^2$ ms scale. \ourmethod achieves even better efficiency, staying within the $10^2$ ms range—lower than Spexpormer and slightly below GT. This efficiency comes from Mamba’s linear-time global modeling, which captures global context without costly attention or sampling. These results highlight Mamba’s advantage in modeling high-order dependencies while maintaining overall efficiency.

\subsection{Ablation Study}
To verify the contribution of each module, we include ablation studies in above experiments. Based on the results, we analyze the contribution of the two modules to DMbaGCN.

\noindent$\rhd$ \textbf{Impact of GCAMba}:
As shown in Table~\ref{tab:performance_com}, the ablation version DMbaGCN w/o GCAMba consistently shows a performance drop across all datasets (e.g., 89.67\% to 90.49\% on Pubmed), indicating that the global information captured by GCAMba contributes to improving the quality of node representations. This effect becomes even more pronounced in Table~\ref{tab:performance_com_under_dif_lay}, where GCAMba enhances performance at all depth settings, with the improvement being especially significant in deeper networks (e.g., 94.28\% to 95.61\% on Photo under 16 layers).
In addition, Figure~\ref{fig:time_memory_cost} shows that DMbaGCN w/o GCAM has memory and time consumption comparable to GCN. Despite this, GCAMba remains significantly more efficient than Transformer-based methods, highlighting Mamba’s ability to capture global context with low computational and memory overhead.

\noindent$\rhd$ \textbf{Impact of LSEMba}:
The ablation version DMbaGCN w/o LSEM relies solely on GCAMba to aggregate global information. As shown in Table~\ref{tab:performance_com}, its performance is clearly lower than the full model across all datasets, indicating that local information is crucial for effective graph representation learning. DMbaGCN struggles to capture multi-hop structure and node-specific representation evolution without LSEMba, reducing expressiveness. This suggests that global context alone is insufficient and mainly serves to enhance discriminability when combined with strong local features.
\section{Conclusion}
This work presents a Dual Mamba-enhanced Graph Convolutional Network (DMbaGCN), which addresses the over-smoothing issue in deep Graph Neural Networks (GNNs). Over-smoothing is a significant challenge where node representations become indistinguishable due to excessive message passing. DMbaGCN tackles this problem by integrating a selective state-space model named Mamba into the GNN framework in two distinct ways. First, it uses the Local State-Evolution Mamba (LSEMba) to manage the evolution of node representations. Second, it employs the Global Context-Aware Mamba (GCAMba) to capture global dependencies.
Extensive experiments on various datasets shows the superiority of DMbaGCN over deep GNN models and graph transformers, both in terms of classification performance and efficiency.

\section{Acknowledgments}
This work was supported by a grant from the National Natural Science Foundation of China under grants (No.62372211,  62272191), and the Science and Technology Development Program of Jilin Province (No.20250102216JC).

\bibliography{aaai2026}

@article{fu2024vcr,
  title={Vcr-graphormer: A mini-batch graph transformer via virtual connections},
  author={Fu, Dongqi and Hua, Zhigang and Xie, Yan and Fang, Jin and Zhang, Si and Sancak, Kaan and Wu, Hao and Malevich, Andrey and He, Jingrui and Long, Bo},
  journal={arXiv preprint arXiv:2403.16030},
  year={2024}
}

@article{kipf2016semi,
  title={Semi-supervised classification with graph convolutional networks},
  author={Kipf, Thomas N and Welling, Max},
  journal={arXiv preprint arXiv:1609.02907},
  year={2016}
}

@article{velivckovic2017graph,
  title={Graph attention networks},
  author={Veli{\v{c}}kovi{\'c}, Petar and Cucurull, Guillem and Casanova, Arantxa and Romero, Adriana and Lio, Pietro and Bengio, Yoshua},
  journal={arXiv preprint arXiv:1710.10903},
  year={2017}
}

@inproceedings{wu2019simplifying,
  title={Simplifying graph convolutional networks},
  author={Wu, Felix and Souza, Amauri and Zhang, Tianyi and Fifty, Christopher and Yu, Tao and Weinberger, Kilian},
  booktitle={International conference on machine learning},
  pages={6861--6871},
  year={2019},
  organization={PMLR}
}

@article{gasteiger2018predict,
  title={Predict then propagate: Graph neural networks meet personalized pagerank},
  author={Gasteiger, Johannes and Bojchevski, Aleksandar and G{\"u}nnemann, Stephan},
  journal={arXiv preprint arXiv:1810.05997},
  year={2018}
}

@inproceedings{chen2020simple,
  title={Simple and deep graph convolutional networks},
  author={Chen, Ming and Wei, Zhewei and Huang, Zengfeng and Ding, Bolin and Li, Yaliang},
  booktitle={International conference on machine learning},
  pages={1725--1735},
  year={2020},
  organization={PMLR}
}

@article{chien2020adaptive,
  title={Adaptive universal generalized pagerank graph neural network},
  author={Chien, Eli and Peng, Jianhao and Li, Pan and Milenkovic, Olgica},
  journal={arXiv preprint arXiv:2006.07988},
  year={2020}
}

@inproceedings{zhu2021simple,
  title={Simple spectral graph convolution},
  author={Zhu, Hao and Koniusz, Piotr},
  booktitle={International conference on learning representations},
  year={2021}
}

@article{dwivedi2020generalization,
  title={A generalization of transformer networks to graphs},
  author={Dwivedi, Vijay Prakash and Bresson, Xavier},
  journal={arXiv preprint arXiv:2012.09699},
  year={2020}
}

@article{ying2021transformers,
  title={Do transformers really perform badly for graph representation?},
  author={Ying, Chengxuan and Cai, Tianle and Luo, Shengjie and Zheng, Shuxin and Ke, Guolin and He, Di and Shen, Yanming and Liu, Tie-Yan},
  journal={Advances in neural information processing systems},
  volume={34},
  pages={28877--28888},
  year={2021}
}

@article{kreuzer2021rethinking,
  title={Rethinking graph transformers with spectral attention},
  author={Kreuzer, Devin and Beaini, Dominique and Hamilton, Will and L{\'e}tourneau, Vincent and Tossou, Prudencio},
  journal={Advances in Neural Information Processing Systems},
  volume={34},
  pages={21618--21629},
  year={2021}
}

@article{rampavsek2022recipe,
  title={Recipe for a general, powerful, scalable graph transformer},
  author={Ramp{\'a}{\v{s}}ek, Ladislav and Galkin, Michael and Dwivedi, Vijay Prakash and Luu, Anh Tuan and Wolf, Guy and Beaini, Dominique},
  journal={Advances in Neural Information Processing Systems},
  volume={35},
  pages={14501--14515},
  year={2022}
}

@article{shirzad2024even,
  title={Even sparser graph transformers},
  author={Shirzad, Hamed and Lin, Honghao and Venkatachalam, Balaji and Velingker, Ameya and Woodruff, David P and Sutherland, Danica J},
  journal={Advances in Neural Information Processing Systems},
  volume={37},
  pages={71277--71305},
  year={2024}
}

@article{he2025mamba,
  title={Mamba-based graph convolutional networks: Tackling over-smoothing with selective state space},
  author={He, Xin and Wang, Yili and Fan, Wenqi and Shen, Xu and Juan, Xin and Miao, Rui and Wang, Xin},
  journal={arXiv preprint arXiv:2501.15461},
  year={2025}
}

@inproceedings{li2018deeper,
  title={Deeper insights into graph convolutional networks for semi-supervised learning},
  author={Li, Qimai and Han, Zhichao and Wu, Xiao-Ming},
  booktitle={Proceedings of the AAAI conference on artificial intelligence},
  volume={32},
  number={1},
  year={2018}
}

@inproceedings{liu2024enhancing,
  title={Enhancing recommendation systems with GNNs and addressing over-smoothing},
  author={Liu, Wenyi and Zhang, Ziqi and Li, Xinshi and Hu, Jiacheng and Luo, Yuanshuai and Du, Junliang},
  booktitle={2024 4th International Conference on Electronic Information Engineering and Computer Communication (EIECC)},
  pages={1184--1189},
  year={2024},
  organization={IEEE}
}

@article{wu2023demystifying,
  title={Demystifying oversmoothing in attention-based graph neural networks},
  author={Wu, Xinyi and Ajorlou, Amir and Wu, Zihui and Jadbabaie, Ali},
  journal={Advances in Neural Information Processing Systems},
  volume={36},
  pages={35084--35106},
  year={2023}
}

@article{chen2025adedgedrop,
  title={Adedgedrop: Adversarial edge dropping for robust graph neural networks},
  author={Chen, Zhaoliang and Wu, Zhihao and Sadikaj, Ylli and Plant, Claudia and Dai, Hong-Ning and Wang, Shiping and Cheung, Yiu-Ming and Guo, Wenzhong},
  journal={IEEE Transactions on Knowledge and Data Engineering},
  year={2025},
  publisher={IEEE}
}

@inproceedings{chen2024sigformer,
  title={SIGformer: Sign-aware graph transformer for recommendation},
  author={Chen, Sirui and Chen, Jiawei and Zhou, Sheng and Wang, Bohao and Han, Shen and Su, Chanfei and Yuan, Yuqing and Wang, Can},
  booktitle={Proceedings of the 47th international ACM SIGIR conference on research and development in information retrieval},
  pages={1274--1284},
  year={2024}
}

@inproceedings{lin2024gramformer,
  title={Gramformer: Learning crowd counting via graph-modulated transformer},
  author={Lin, Hui and Ma, Zhiheng and Hong, Xiaopeng and Shangguan, Qinnan and Meng, Deyu},
  booktitle={Proceedings of the AAAI Conference on Artificial Intelligence},
  volume={38},
  number={4},
  pages={3395--3403},
  year={2024}
}

@article{han2024demystify,
  title={Demystify mamba in vision: A linear attention perspective},
  author={Han, Dongchen and Wang, Ziyi and Xia, Zhuofan and Han, Yizeng and Pu, Yifan and Ge, Chunjiang and Song, Jun and Song, Shiji and Zheng, Bo and Huang, Gao},
  journal={Advances in neural information processing systems},
  volume={37},
  pages={127181--127203},
  year={2024}
}

@article{patro2024simba,
  title={Simba: Simplified mamba-based architecture for vision and multivariate time series},
  author={Patro, Badri N and Agneeswaran, Vijay S},
  journal={arXiv preprint arXiv:2403.15360},
  year={2024}
}

@article{wang2024graph,
  title={Graph-mamba: Towards long-range graph sequence modeling with selective state spaces},
  author={Wang, Chloe and Tsepa, Oleksii and Ma, Jun and Wang, Bo},
  journal={arXiv preprint arXiv:2402.00789},
  year={2024}
}

@inproceedings{behrouz2024graph,
  title={Graph mamba: Towards learning on graphs with state space models},
  author={Behrouz, Ali and Hashemi, Farnoosh},
  booktitle={Proceedings of the 30th ACM SIGKDD Conference on Knowledge Discovery and Data Mining},
  pages={119--130},
  year={2024}
}

@inproceedings{ding2024recurrent,
  title={Recurrent Distance Filtering for Graph Representation Learning},
  author={Ding, Yuhui and Orvieto, Antonio and He, Bobby and Hofmann, Thomas},
  booktitle={Forty-first International Conference on Machine Learning},
  year={2024}
}

@inproceedings{yang2024graph,
  title={Graph bottlenecked social recommendation},
  author={Yang, Yonghui and Wu, Le and Wang, Zihan and He, Zhuangzhuang and Hong, Richang and Wang, Meng},
  booktitle={Proceedings of the 30th ACM SIGKDD Conference on Knowledge Discovery and Data Mining},
  pages={3853--3862},
  year={2024}
}

@article{khan2025heterogeneous,
  title={Heterogeneous hypergraph neural network for social recommendation using attention network},
  author={Khan, Bilal and Wu, Jia and Yang, Jian and Ma, Xiaoxiao},
  journal={ACM Transactions on Recommender Systems},
  volume={3},
  number={3},
  pages={1--22},
  year={2025},
  publisher={ACM New York, NY}
}

@article{sypetkowski2024scalability,
  title={On the scalability of gnns for molecular graphs},
  author={Sypetkowski, Maciej and Wenkel, Frederik and Poursafaei, Farimah and Dickson, Nia and Suri, Karush and Fradkin, Philip and Beaini, Dominique},
  journal={Advances in Neural Information Processing Systems},
  volume={37},
  pages={19870--19906},
  year={2024}
}

@inproceedings{wang2023vqa,
  title={Vqa-gnn: Reasoning with multimodal knowledge via graph neural networks for visual question answering},
  author={Wang, Yanan and Yasunaga, Michihiro and Ren, Hongyu and Wada, Shinya and Leskovec, Jure},
  booktitle={Proceedings of the IEEE/CVF international conference on computer vision},
  pages={21582--21592},
  year={2023}
}

@article{waleffe2024empirical,
  title={An empirical study of mamba-based language models},
  author={Waleffe, Roger and Byeon, Wonmin and Riach, Duncan and Norick, Brandon and Korthikanti, Vijay and Dao, Tri and Gu, Albert and Hatamizadeh, Ali and Singh, Sudhakar and Narayanan, Deepak and others},
  journal={arXiv preprint arXiv:2406.07887},
  year={2024}
}

@inproceedings{wang2025mamba,
  title={Mamba-Reg: Vision Mamba Also Needs Registers},
  author={Wang, Feng and Wang, Jiahao and Ren, Sucheng and Wei, Guoyizhe and Mei, Jieru and Shao, Wei and Zhou, Yuyin and Yuille, Alan and Xie, Cihang},
  booktitle={Proceedings of the Computer Vision and Pattern Recognition Conference},
  pages={14944--14953},
  year={2025}
}

@inproceedings{hatamizadeh2025mambavision,
  title={Mambavision: A hybrid mamba-transformer vision backbone},
  author={Hatamizadeh, Ali and Kautz, Jan},
  booktitle={Proceedings of the Computer Vision and Pattern Recognition Conference},
  pages={25261--25270},
  year={2025}
}

@inproceedings{wang2025vggt,
  title={Vggt: Visual geometry grounded transformer},
  author={Wang, Jianyuan and Chen, Minghao and Karaev, Nikita and Vedaldi, Andrea and Rupprecht, Christian and Novotny, David},
  booktitle={Proceedings of the Computer Vision and Pattern Recognition Conference},
  pages={5294--5306},
  year={2025}
}

@article{dao2024transformers,
  title={Transformers are ssms: Generalized models and efficient algorithms through structured state space duality},
  author={Dao, Tri and Gu, Albert},
  journal={arXiv preprint arXiv:2405.21060},
  year={2024}
}

@article{gu2020hippo,
  title={Hippo: Recurrent memory with optimal polynomial projections},
  author={Gu, Albert and Dao, Tri and Ermon, Stefano and Rudra, Atri and R{\'e}, Christopher},
  journal={Advances in neural information processing systems},
  volume={33},
  pages={1474--1487},
  year={2020}
}

@inproceedings{song2024breaking,
  title={Breaking the Bottleneck on Graphs with Structured State Spaces},
  author={Song, Yunchong and Huang, Siyuan and Cai, Jiacheng and Wang, Xinbing and Zhou, Chenghu and Lin, Zhouhan},
  booktitle={Proceedings of the 33rd ACM International Conference on Information and Knowledge Management},
  pages={2138--2147},
  year={2024}
}

@article{he2021bernnet,
  title={Bernnet: Learning arbitrary graph spectral filters via bernstein approximation},
  author={He, Mingguo and Wei, Zhewei and Xu, Hongteng and others},
  journal={Advances in neural information processing systems},
  volume={34},
  pages={14239--14251},
  year={2021}
}

@article{ali2024hidden,
  title={The hidden attention of mamba models},
  author={Ali, Ameen and Zimerman, Itamar and Wolf, Lior},
  journal={arXiv preprint arXiv:2403.01590},
  year={2024}
}

@article{shen2024graph,
  title={Graph rewiring and preprocessing for graph neural networks based on effective resistance},
  author={Shen, Xu and Lio, Pietro and Yang, Lintao and Yuan, Ru and Zhang, Yuyang and Peng, Chengbin},
  journal={IEEE Transactions on Knowledge and Data Engineering},
  volume={36},
  number={11},
  pages={6330--6343},
  year={2024},
  publisher={IEEE}
}

@article{chen2024leveraging,
  title={Leveraging contrastive learning for enhanced node representations in tokenized graph transformers},
  author={Chen, Jinsong and Liu, Hanpeng and Hopcroft, John and He, Kun},
  journal={Advances in Neural Information Processing Systems},
  volume={37},
  pages={85824--85845},
  year={2024}
}

@inproceedings{roth2024rank,
  title={Rank collapse causes over-smoothing and over-correlation in graph neural networks},
  author={Roth, Andreas and Liebig, Thomas},
  booktitle={Learning on Graphs Conference},
  pages={35--1},
  year={2024},
  organization={PMLR}
}

@article{velivckovic2023everything,
  title={Everything is connected: Graph neural networks},
  author={Veli{\v{c}}kovi{\'c}, Petar},
  journal={Current Opinion in Structural Biology},
  volume={79},
  pages={102538},
  year={2023},
  publisher={Elsevier}
}

@article{gao2024matten,
  title={Matten: Video generation with mamba-attention},
  author={Gao, Yu and Huang, Jiancheng and Sun, Xiaopeng and Jie, Zequn and Zhong, Yujie and Ma, Lin},
  journal={arXiv preprint arXiv:2405.03025},
  year={2024}
}

@article{choi2024topology,
  title={Topology-informed graph transformer},
  author={Choi, Yun Young and Park, Sun Woo and Lee, Minho and Woo, Youngho},
  journal={arXiv preprint arXiv:2402.02005},
  year={2024}
}

@article{wang2025adagcl+,
  title={AdaGCL+: An Adaptive Subgraph Contrastive Learning Towards Tackling Topological Bias},
  author={Wang, Yili and Liu, Yaohua and Liu, Ninghao and Miao, Rui and Wang, Ying and Wang, Xin},
  journal={IEEE Transactions on Pattern Analysis and Machine Intelligence},
  year={2025},
  publisher={IEEE}
}

@InProceedings{miao2024rethinking,
  title={Rethinking Independent Cross-Entropy Loss For Graph-Structured Data},
  author={Miao, Rui and Zhou, Kaixiong and Wang, Yili and Liu, Ninghao and Wang, Ying and Wang, Xin},
  booktitle={Proceedings of the 41st International Conference on Machine Learning},
  pages={35570--35589},
  year={2024},
  volume={235},
  publisher={PMLR}
}

@inproceedings{GRASS,
  author       = {Liang Yang and
                  Yukun Cai and
                  Hui Ning and
                  Jiaming Zhuo and
                  Di Jin and
                  Ziyi Ma and
                  Yuanfang Guo and
                  Chuan Wang and
                  Zhen Wang},
  title        = {Universal Graph Self-Contrastive Learning},
  booktitle    = {IJCAI},
  pages        = {3534--3542},
  year={2025},
}

@inproceedings{BQN,
  title={Do We Really Need Message Passing in Brain Network Modeling?},
  author={Yang, Liang and Liu, Yuwei and Zhuo, Jiaming and Jin, Di and Wang, Chuan and Wang, Zhen and Cao, Xiaochun},
  booktitle={ICML}, 
  year={2025},
}

@inproceedings{FuDHL0C25,
  author       = {Lele Fu and
                  Bowen Deng and
                  Sheng Huang and
                  Tianchi Liao and
                  Chuanfu Zhang and
                  Chuan Chen},
  title        = {Learn from Global Rather Than Local: Consistent Context-Aware Representation
                  Learning for Multi-View Graph Clustering},
  booktitle    = {Proceedings of the Thirty-Fourth International Joint Conference on
                  Artificial Intelligence, {IJCAI} 2025, Montreal, Canada, August 16-22,
                  2025},
  pages        = {5145--5153},
  year         = {2025}
}

@inproceedings{shen2024optimizing,
  title={Optimizing ood detection in molecular graphs: A novel approach with diffusion models},
  author={Shen, Xu and Wang, Yili and Zhou, Kaixiong and Pan, Shirui and Wang, Xin},
  booktitle={Proceedings of the 30th ACM SIGKDD Conference on Knowledge Discovery and Data Mining},
  pages={2640--2650},
  year={2024}
}

@article{wang2024unifying,
  title={Unifying unsupervised graph-level anomaly detection and out-of-distribution detection: A benchmark},
  author={Wang, Yili and Liu, Yixin and Shen, Xu and Li, Chenyu and Ding, Kaize and Miao, Rui and Wang, Ying and Pan, Shirui and Wang, Xin},
  journal={arXiv preprint arXiv:2406.15523},
  year={2024}
}

@article{shen2025raising,
  title={Raising the bar in graph ood generalization: Invariant learning beyond explicit environment modeling},
  author={Shen, Xu and Liu, Yixin and Wang, Yili and Miao, Rui and Dai, Yiwei and Pan, Shirui and Chang, Yi and Wang, Xin},
  journal={arXiv preprint arXiv:2502.10706},
  year={2025}
}

@article{shen2025understanding,
  title={Understanding the Information Propagation Effects of Communication Topologies in LLM-based Multi-Agent Systems},
  author={Shen, Xu and Liu, Yixin and Dai, Yiwei and Wang, Yili and Miao, Rui and Tan, Yue and Pan, Shirui and Wang, Xin},
  journal={arXiv preprint arXiv:2505.23352},
  year={2025}
}
\appendix


\section{A Experiments Detail}

\subsection{A.1 Dataset}\label{appendix:dataset}
The statistics are listed in Table~\ref{tab:datasets}.
Pubmed and CoraFull are citation graph datasets, Computers and Photo are web graph datasets, CS and Physics are co-authorship graph datasets.
We utilize the feature vectors, class labels, and 10 random splits as proposed by~\cite{chen2020simple} for citation graph datasets and heterogeneous graph datasets. 
We use the feature vectors, class labels, and 10 random splits as detailed in~\cite{he2021bernnet} for web graph datasets. All experiments are performed on a system with an Intel(R) Xeon(R) Gold 5120 CPU and an NVIDIA L40 48G GPU.
\begin{itemize}
    \item \textbf{Pubmed \& CoraFull}: These two datasets are benchmark citation network datasets, where nodes represent papers and edges denote citation relationships.
    \item \textbf{Computers \& Photo}: These two datasets are commonly used node classification datasets from the Amazon co-purchase graph. In these datasets, nodes represent goods, and edges indicate that two goods are frequently bought together. 
    \item \textbf{CS \& Physics}: These two datasets are widely used node classification benchmarks derived from the Microsoft Academic Graph. Nodes represent authors, and edges indicate co-authorship; two authors are connected if they have written a paper together. Each author's feature vector is constructed by aggregating the keywords of their published papers using a bag-of-words approach.
\end{itemize}

\begin{table}[hb]
\renewcommand{\arraystretch}{1.2}
\centering
\scalebox{1}{
\begin{tabular}{c|cccc}
\toprule
\multicolumn{1}{c|}{\textbf{Datasets}}  & \multicolumn{1}{c}{\textbf{Nodes}}& \multicolumn{1}{c}{\textbf{Edges}}& \multicolumn{1}{c}{\textbf{Features}}& \multicolumn{1}{c}{\textbf{Classes}}\\
\midrule
 \multicolumn{1}{c|}{\textbf{Pubmed}} &  19,717  & 88,651  &  500  &   3  \\
\multicolumn{1}{c|}{\textbf{CoraFull}}& 19,793   &   128,824&  8,710  &  70   \\
\multicolumn{1}{c|}{\textbf{Computer}}& 13,752   &   491,722 &  767  &  10   \\
\multicolumn{1}{c|}{\textbf{Photo}}& 7,650   &  238,163 &   745 &  8   \\
\multicolumn{1}{c|}{\textbf{CS}}& 18,333   &  163,788 &   6,805 &  15   \\
\multicolumn{1}{c|}{\textbf{Physics}}& 34,493   &  495,924 &  8,415 & 15   \\
\bottomrule
\end{tabular}}
\caption{Dataset statistics.} \label{tab:datasets}
\end{table}


    

\begin{table*}[t]
\renewcommand{\arraystretch}{1.2}
\centering
\scalebox{0.78}{
\begin{tabular}{c|ccccc|ccccc}
\toprule
\multicolumn{1}{c|}{\textbf{Layers}}  & \multicolumn{1}{c}{\textbf{2}}& \multicolumn{1}{c}{\textbf{4}}& \multicolumn{1}{c}{\textbf{8}}& \multicolumn{1}{c}{\textbf{16}}& \multicolumn{1}{c}{\textbf{32}}& \multicolumn{1}{|c}{\textbf{2}}& \multicolumn{1}{c}{\textbf{4}}& \multicolumn{1}{c}{\textbf{8}}& \multicolumn{1}{c}{\textbf{16}}& \multicolumn{1}{c}{\textbf{32}}\\
\midrule

\textbf{Dataset}&\multicolumn{5}{c|}{\textbf{CS}}&\multicolumn{5}{c}{\textbf{CoraFull}} \\
\cmidrule{1-11}
\multirow{1}{*}{\textbf{GCN}}  & \textbf{93.75±0.26} & \underline{92.31±0.24} & 89.50±0.25 & 49.06±2.09& 22.66±0.36& \textbf{70.69±0.37} & \underline{68.54±0.34} & 66.10±0.27 & 43.15±4.80& 8.71±0.34\\
\multirow{1}{*}{\textbf{SGC}}& \textbf{93.44±0.18}  & \underline{92.37±0.27} & 91.50±0.30 & 90.26±0.32 & 87.94±0.29 & 
\textbf{70.04±0.25} & \underline{68.67±0.34} & 67.40±0.35 & 65.59±0.29 & 59.02±0.86\\ \midrule
                \multirow{1}{*}{\textbf{APPNP}}& 95.64±0.19 & 95.65±0.19 & \underline{95.64±0.18} & \textbf{95.65±0.17} & 95.64±0.19 & 69.31±0.26 & 69.22±0.47 & 69.36±0.39 & \textbf{69.37±0.35} & \underline{69.36±0.37}\\
                \multirow{1}{*}{\textbf{GCNII}}
& 95.42±0.13 & 95.42±0.15 & \underline{95.43±0.14} & 95.43±0.18 & \textbf{95.46±0.14} & 69.20±0.47 & 69.94±0.44 & 69.92±0.54 & \underline{71.37±0.41} & \textbf{72.23±0.50}\\
\multirow{1}{*}{\textbf{GPRGNN}}
& 95.35±0.16 & \underline{95.42±0.19}&  \textbf{95.49±0.19}& 95.34±0.16 & 95.41±0.16& \textbf{71.16±0.50} & \underline{70.90±0.52}&  70.54±0.58& 70.68±0.48 & 69.42±0.49\\ 
\multirow{1}{*}{\textbf{SSGC}}& \textbf{93.99±0.23}& \underline{93.81±0.24} & 93.51±0.28 & 93.45±0.28 & 93.70±0.24 & 70.43±0.39& 70.36±0.39 & 70.28±0.26 & \textbf{70.51±0.25} & \underline{70.44±0.42}\\  \cmidrule{1-11}
\multirow{2}{*}{\textbf{\makecell[c]{\ourmethod \\w/o GMPM}}}& \multirow{2}{*}{\underline{95.24±0.25}}& \multirow{2}{*}{\textbf{95.32±0.15}} & \multirow{2}{*}{95.10±0.19} & \multirow{2}{*}{95.07±0.19} & \multirow{2}{*}{95.12±0.23}& \multirow{2}{*}{71.37±0.41}& \multirow{2}{*}{\underline{71.42±0.50}} & \multirow{2}{*}{\textbf{71.51±0.37}} & \multirow{2}{*}{70.87±0.35} & \multirow{2}{*}{71.02±0.43}\\ &&&&&\multicolumn{1}{c|}{} \\
\multirow{2}{*}{\textbf{\makecell[c]{\ourmethod \\(ours)}}}& \multirow{2}{*}{95.66±0.31}& \multirow{2}{*}{\textbf{96.00±0.21}} & \multirow{2}{*}{\underline{95.73±0.41}} &\multirow{2}{*}{95.55±0.25} & \multirow{2}{*}{95.51±0.20}& \multirow{2}{*}{72.07±0.50}& \multirow{2}{*}{\underline{72.08±0.40}} & \multirow{2}{*}{\textbf{72.26±0.20}} &\multirow{2}{*}{71.45±0.27} & \multirow{2}{*}{71.68±0.31}\\ &&&&&\multicolumn{1}{c|}{} \\ 
\midrule

\textbf{Dataset}&\multicolumn{5}{c|}{\textbf{Computer}}&\multicolumn{5}{c}{\textbf{Physics}} \\
\cmidrule{1-11}
\multirow{1}{*}{\textbf{GCN}}  &  \textbf{91.07±0.39}
 & \underline{88.46±0.28} & 80.48±0.43 & 63.46±1.61 & 38.47±0.41 & \textbf{96.34±0.21} & \underline{95.92±0.15} & 95.52±0.14 & 94.09±0.17 & 65.75±0.13\\
\multirow{1}{*}{\textbf{SGC}}& \textbf{91.62±0.35}  & \underline{89.69±0.30} & 84.43±0.46 & 79.18±0.55 & 75.53±0.66 & 
 \textbf{96.43±0.20} & \underline{96.06±0.14} & 95.64±0.14 & 95.03±0.13 & 93.93±0.15\\ \midrule
                \multirow{1}{*}{\textbf{APPNP}}&
\textbf{89.51±0.31}  & \underline{89.51±0.36} & 89.41±0.41 & 89.44±0.28 & 89.44±0.33 & 96.99±0.17 & \textbf{97.01±0.17} & \underline{96.99±0.14} & 96.99±0.17 & 96.99±0.16\\
                \multirow{1}{*}{\textbf{GCNII}}
& 83.64±0.68  & 83.83±0.98 & \underline{84.54±0.59} & \textbf{84.71±0.40} & 84.03±0.59 & \underline{97.05±0.15} & \textbf{97.09±0.13} & 96.94±0.15 & 96.95±0.12 & 96.94±0.14\\
\multirow{1}{*}{\textbf{GPRGNN}}
& 91.37±0.28  & 91.44±0.35 & 91.38±0.47 & \textbf{91.80±0.35} & \underline{91.73±0.26}& \textbf{97.05±0.13} & \underline{97.02±0.14}& 97.01±0.14& 96.94±0.13 & 96.69±0.12\\ 
\multirow{1}{*}{\textbf{SSGC}}& \underline{91.96±0.33}  & \textbf{91.98±0.36} & 91.39±0.49 & 90.52±0.45 & 89.21±0.48 & \textbf{96.60±0.15}& 96.54±0.17 & \underline{96.55±0.14} & 96.50±0.22 & 96.50±0.21\\  \cmidrule{1-11}
\multirow{2}{*}{\textbf{\makecell[c]{\ourmethod \\w/o GMPM}}}& \multirow{2}{*}{\textbf{91.74±0.34}}  & \multirow{2}{*}{\underline{91.53±0.42}}
& \multirow{2}{*}{90.78±0.29} & \multirow{2}{*}{90.47±0.28} & \multirow{2}{*}{90.08±0.31}& \multirow{2}{*}{\textbf{96.37±0.16}}& \multirow{2}{*}{\underline{96.12±0.15}} & \multirow{2}{*}{96.03±0.14} & \multirow{2}{*}{96.02±0.21} & \multirow{2}{*}{95.87±0.16}\\ &&&&&\multicolumn{1}{c|}{} \\
\multirow{2}{*}{\textbf{\makecell[c]{\ourmethod \\(ours)}}}& \multirow{2}{*}{\textbf{92.49±0.37}}  & \multirow{2}{*}{\underline{92.12±0.34}}
& \multirow{2}{*}{91.52±0.29} & \multirow{2}{*}{91.41±0.23} & \multirow{2}{*}{91.21±0.29}& \multirow{2}{*}{\textbf{97.13±0.14}}& \multirow{2}{*}{\underline{97.04±0.15}} & \multirow{2}{*}{96.95±0.12} &\multirow{2}{*}{97.02±0.13} & \multirow{2}{*}{96.99±0.18}\\ &&&&&\multicolumn{1}{c|}{} \\ 
\bottomrule
\end{tabular}}
\caption{Classification accuracy (\%) comparison under different layer configurations. The best result across different layer configurations is highlighted in \textbf{bold}, and the second-best result is emphasized with an \underline{underline}.} \label{tab:performance_com_under_dif_lay_appendix}
\end{table*}

\begin{figure*}[t!]
\includegraphics[width=0.245\linewidth]{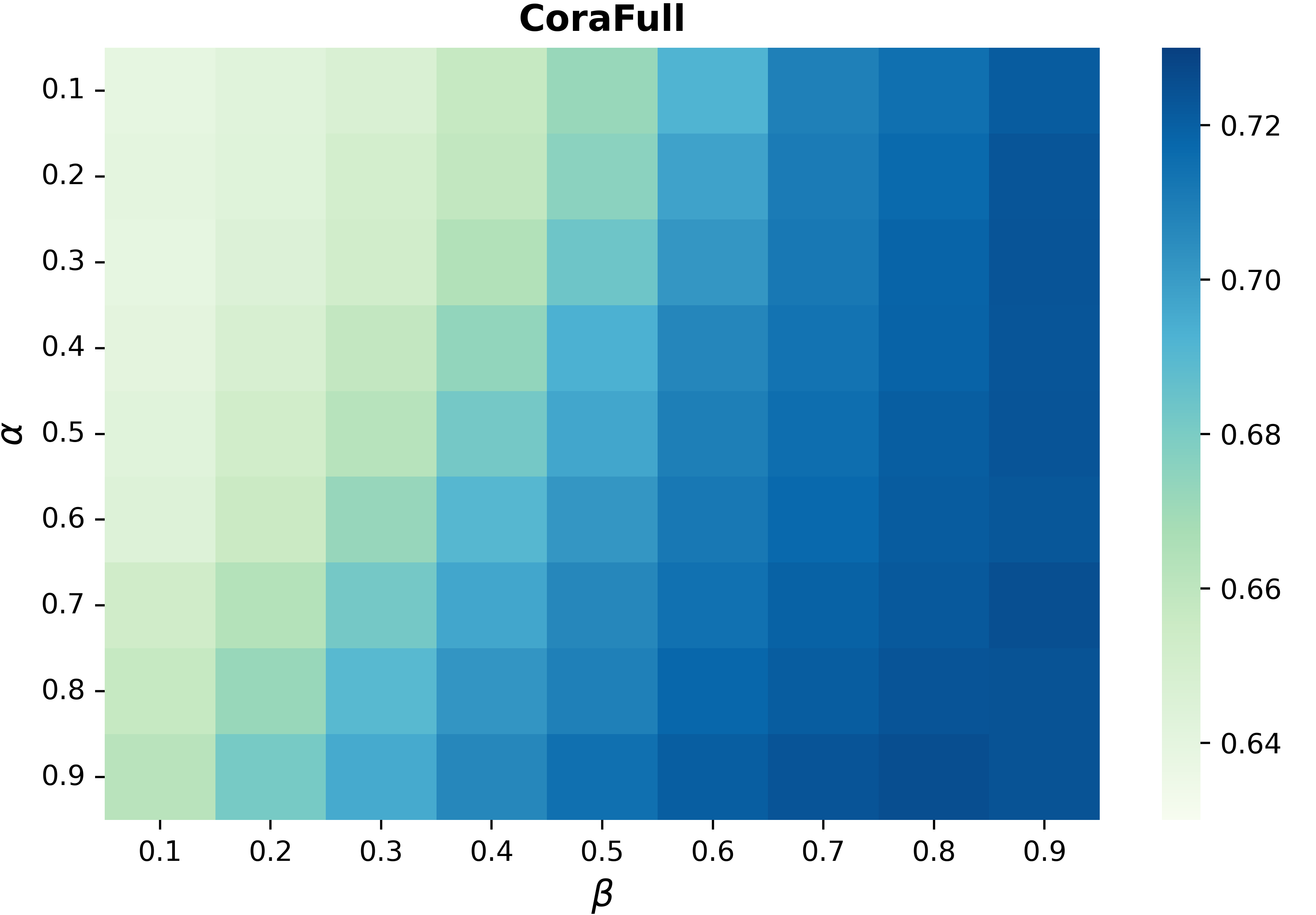}
\includegraphics[width=0.245\linewidth]{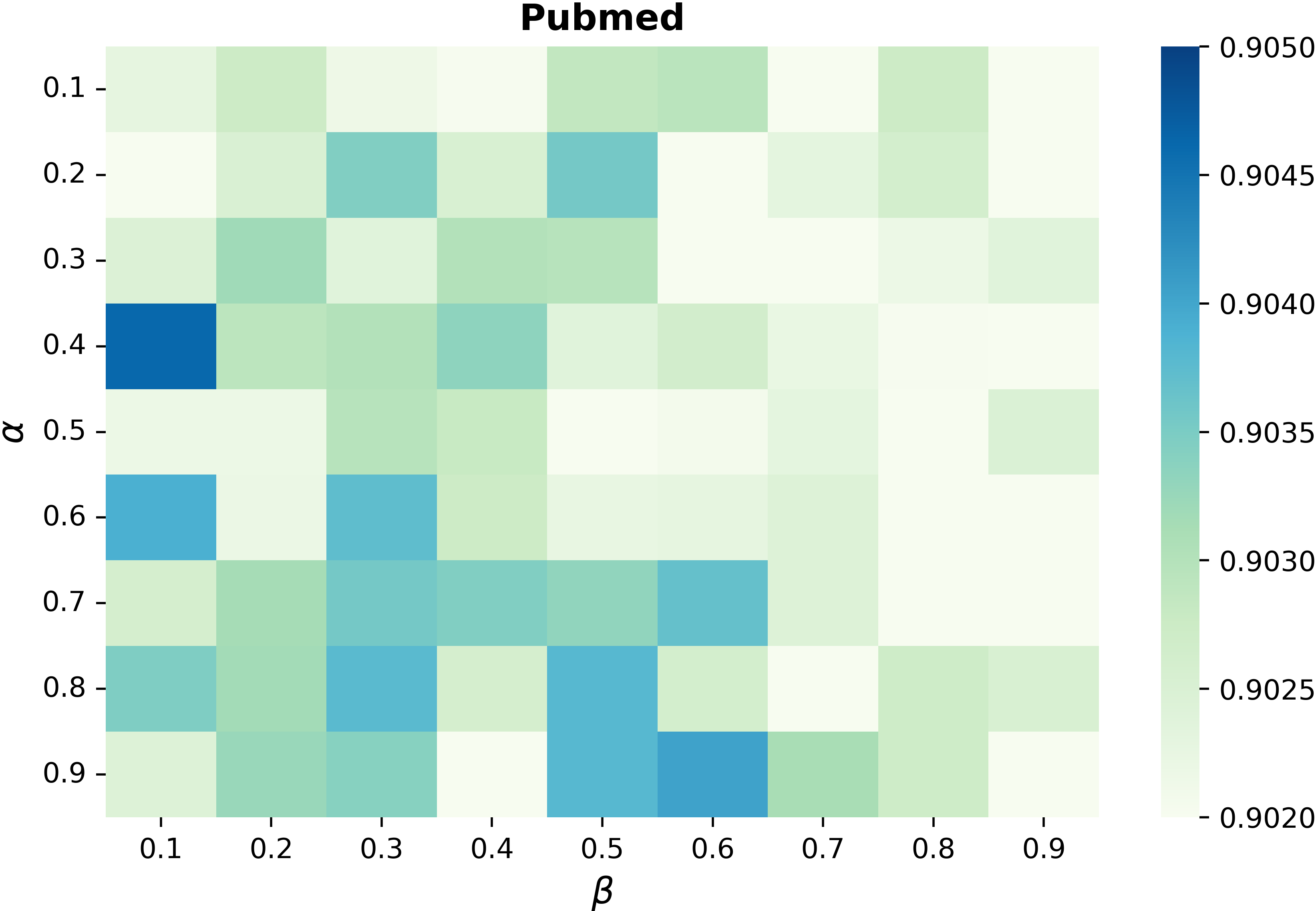}
\includegraphics[width=0.245\linewidth]{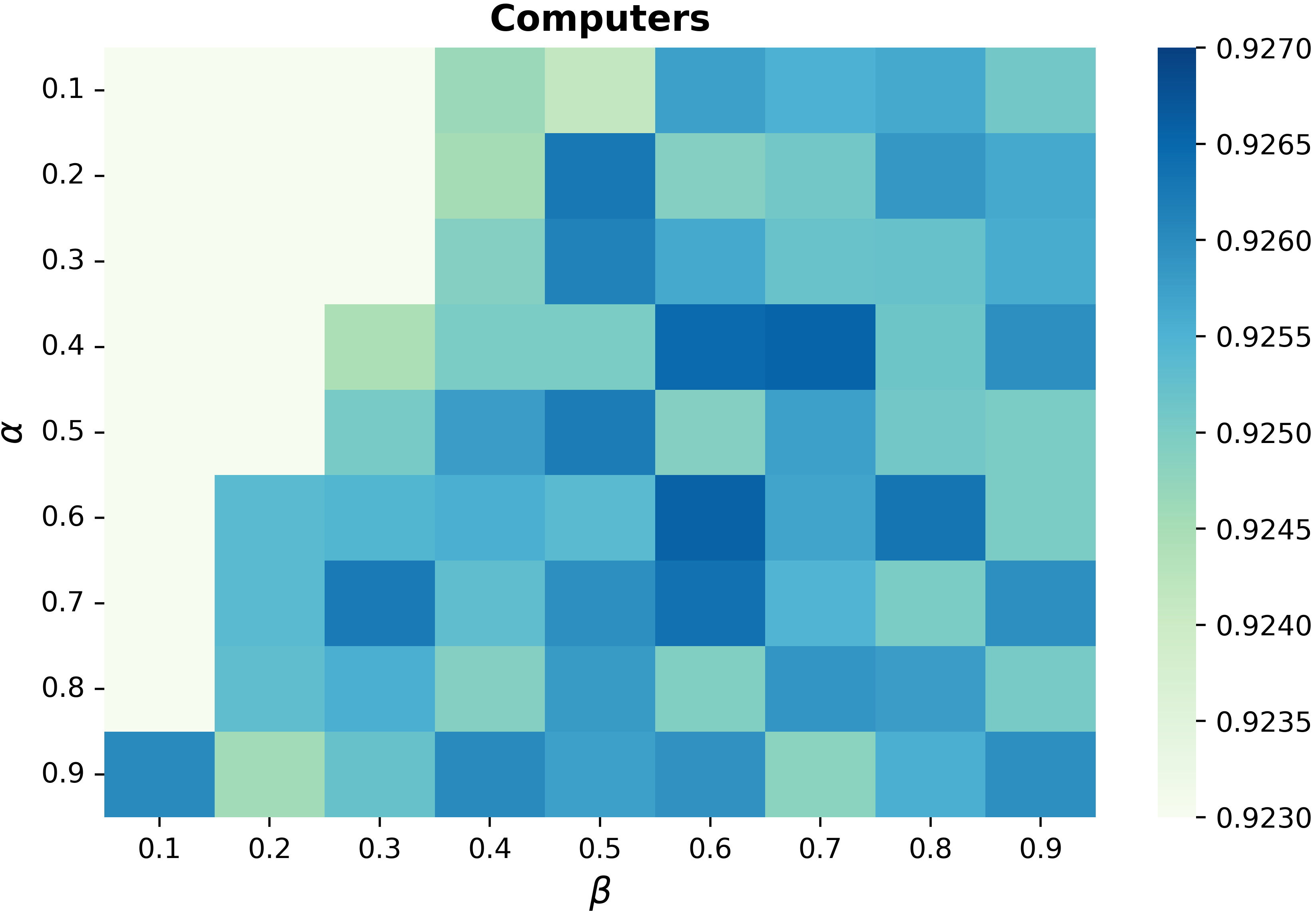}
\includegraphics[width=0.245\linewidth]{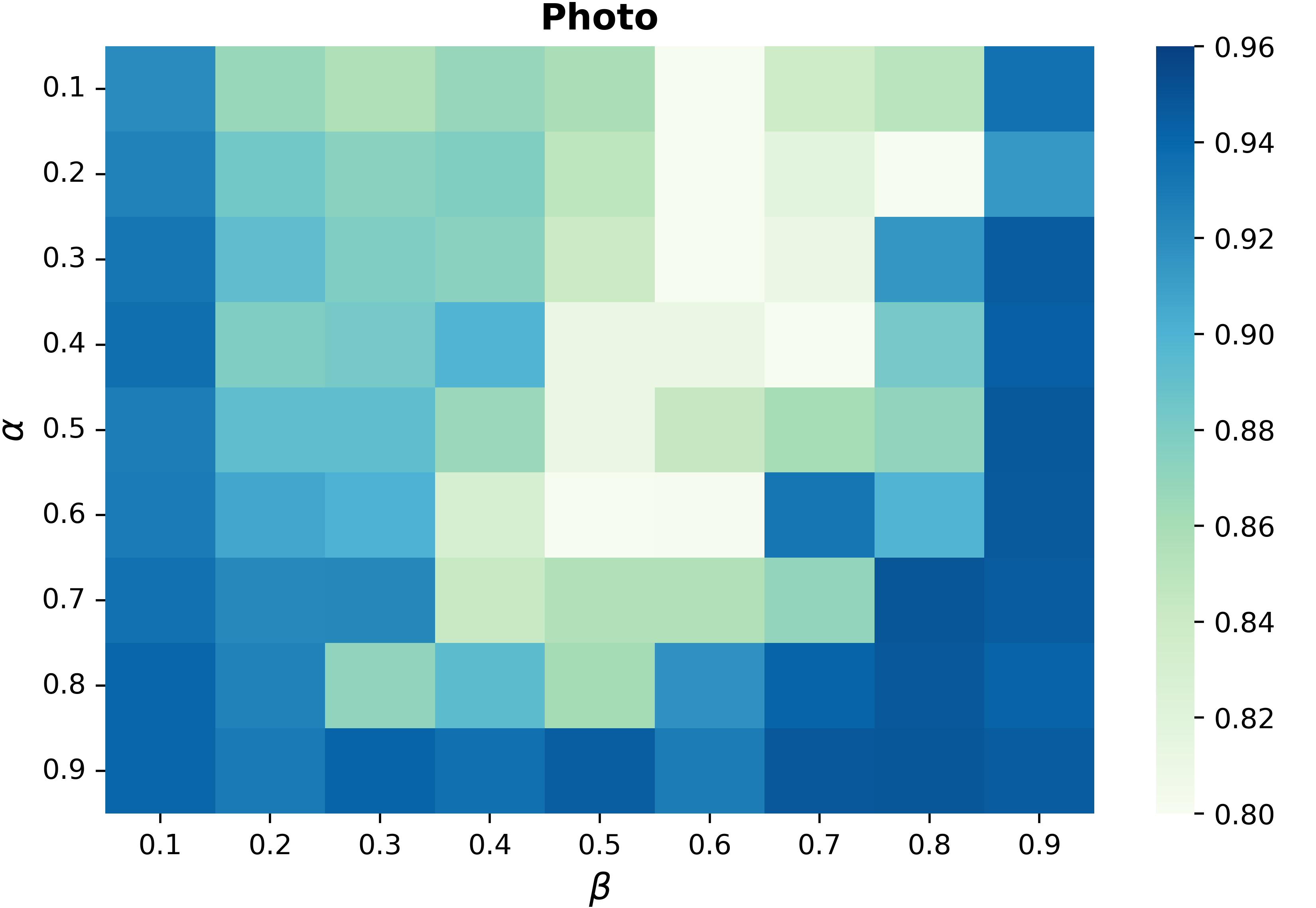}
  \caption{Effect of Hyperparameters $\alpha$ and $\beta$ on Model Performance}
  \label{fig:hperparameters}
\end{figure*}

\subsection{A.2 Baselines}
\begin{itemize}
    \item \textbf{GCN}: This is a foundational graph neural network model that improves stability and scalability by employing a first-order approximation of spectral graph convolutions.
    \item \textbf{GAT}: This graph neural network leverages an attention mechanism to adaptively aggregate information from neighboring nodes, enabling implicit weighting based on the importance of node features within the local neighborhood.
    \item \textbf{SGC}: This method is a simplified variant of GCN that removes nonlinearities and weight matrices between graph convolution layers, resulting in a more efficient and interpretable model.
    \item \textbf{APPNP}: This method enhances the propagation mechanism of GCN by incorporating personalized PageRank, enabling more effective and flexible information diffusion across the graph.
    \item \textbf{GCNII}: This method is a variant of GCN that integrates residual connections and identity mapping to effectively mitigate the over-smoothing problem in deep graph networks.
    \item \textbf{GPRGNN}: This model integrates personalized PageRank into GNNs to better capture node importance, improving performance and mitigating over-smoothing.
    \item \textbf{SSGC}: This method alleviates over-smoothing by applying a modified Markov diffusion kernel that symmetrically scales the aggregation process, effectively preserving node feature distinctiveness in deeper GNN layers.
    \item \textbf{GT}: This is a transformer-based model for graphs that uses neighborhood-aware attention and Laplacian positional encodings to capture structural information.
    \item \textbf{Graphormer}: This method adapts the standard Transformer to graphs by introducing structural encodings for node centrality, spatial distance, and edge features.
    \item \textbf{SAN}: This method introduces spectral-based attention to Graph Transformers by leveraging the eigenspace of the graph Laplacian. This allows the model to capture global structural information more effectively, leading to strong performance across various graph learning tasks.
    \item \textbf{GraphGPS}: This method combines local message passing (like GNNs) with global attention (like Transformers) in a unified framework. By integrating both spatial and global structural information, GPS achieves strong and scalable performance across diverse graph tasks.
    \item \textbf{Spexphormer}: This method designs sparser attention patterns for Graph Transformers using spectral and spatial priors, significantly reducing computation while maintaining strong performance. It enables scalable and efficient training on large graphs.
    \item \textbf{MbaGCN}: This method integrates the Mamba-inspired selective state space mechanism into GNNs to alleviate over-smoothing. By combining message aggregation with adaptive neighborhood filtering, it enables deeper architectures.
\end{itemize}

\section{B Experiment}
\subsection{B.1 Layer-Wise Performance Trends}

As shown in Table~\ref{tab:performance_com_under_dif_lay_appendix}, the classification accuracy (\%) of GNN and deep GNN architectures is reported across multiple depths (2–32 layers) on the CS, CoraFull, Computer and Physics datasets. GCN and SGC exhibit a clear trend of performance degradation as depth increases, likely due to their vulnerability to over-smoothing. In comparison, deeper GNN variants like GCNII demonstrate stronger depth robustness, with accuracy that remains consistent or improves incrementally with additional layers.

Table~\ref{tab:performance_com_under_dif_lay_appendix} shows that our proposed DMbaGCN not only surpasses most baseline methods as the network depth increases but also maintains performance stability. This demonstrates the effectiveness of leveraging the Mamba mechanism to alleviate over-smoothing in deep GNNs. Additionally, the results from the ablation study underscore the critical role of global information in improving node separability.

\subsection{B.2 Hyperparameter Analyses}
To evoluate the impact of hyperparameters on model performance, we conduct experiments on four datasets (CoraFull, Pubmed, Computers and Photo), focusing on the joint effect of $\alpha$ and $\beta$. Specifically, $\beta$ controls the degree of residual connection during global aggregation, while $\alpha$ balances the contribution of local and global information in the final node representation. As shown in the Figure~\ref{fig:hperparameters}, we perform a grid search over $\alpha$ and $\beta$ within the range of $[0.1, 0.2, 0.3, 0.4, 0.5, 0.6, 0.7, 0.8, 0.9]$. 
The results reveal that the optimal combination of $\alpha$ and $\beta$ varies significantly across datasets.
In contrast, Computers and Photo achieve better accuracy under moderate or balanced values of both $\alpha$ and $\beta$, indicating that a proper fusion of local and global features with initial features is essential.
We select the best-performing combination of $\alpha$ and $\beta$ for each dataset as the final hyperparameter setting used in our experiments.

\end{document}